\documentclass[10pt,twocolumn,letterpaper]{article}

\usepackage{iccv}
\usepackage{times}
\usepackage{epsfig}

\usepackage{amsmath}
\usepackage{amssymb}  
\usepackage{mathabx}
\usepackage{footmisc}
\usepackage{subfiles}

\newtheorem{prop}{Proposition}
\DeclareMathOperator{\argmax}{arg\,max}

\usepackage{centernot}
\usepackage{array}
\usepackage{multirow}
\usepackage{amsfonts}
\usepackage{xcolor}
\usepackage[utf8]{inputenc}
\usepackage{comment}
\usepackage{todonotes}
\usepackage{amsfonts, bbm}
\usepackage{xcolor}
\usepackage{tikz}
\usepackage[pagebackref=true,breaklinks=true,letterpaper=true,colorlinks,bookmarks=false]{hyperref}
\usepackage{tabularx}
\usepackage{subcaption} 
\usepackage{xpatch}
\usepackage{lscape}
\usepackage[misc,geometry]{ifsym}
\usepackage{lipsum}
\usepackage[absolute,overlay]{textpos}

\newcount\Comments  
\usepackage{color}
\definecolor{darkgreen}{rgb}{0,0.5,0}
\definecolor{purple}{rgb}{1,0,1}
\newcommand{\kibitz}[2]{\ifnum\Comments=1\textcolor{#1}{#2}\fi}

\usepackage[T1]{fontenc}
\usepackage{graphicx}
\usepackage[absolute,overlay]{textpos}

\usepackage[breaklinks=true,bookmarks=false]{hyperref}

 \iccvfinalcopy 

\def\iccvPaperID{4952} 

\ificcvfinal\fi

\begin{document}

\begin{textblock*}{20cm}(1cm,1cm)
\textcolor{red}{Accepted by the IEEE/CVF International Conference on Computer Vision 2023 (ICCV'23)}
\end{textblock*}

\title{SAFARI: Versatile and Efficient Evaluations for Robustness of Interpretability}


\author{
Wei Huang$^{1,2}$ \qquad Xingyu Zhao$^{2,3}$ \qquad Gaojie Jin$^{2,4}$ \qquad Xiaowei Huang$^{2}$ \\
$^{1}$Purple Mountain Laboratories \quad $^{2}$University of Liverpool \\
$^{3}$WMG, University of Warwick \quad $^{4}$Institute of Software, CAS\\
{\tt\small\{w.huang23,xingyu.zhao,g.jin3,xiaowei.huang\}@liverpool.ac.uk}
}

\maketitle
\ificcvfinal\thispagestyle{empty}\fi

\begin{abstract}

Interpretability of Deep Learning (DL) is a barrier to trustworthy AI. Despite great efforts made by the Explainable AI (XAI) community, explanations lack robustness---indistinguishable input perturbations may lead to different XAI results. Thus, it is vital to assess how robust DL interpretability is, given an XAI method. 
In this paper, we identify several challenges that the state-of-the-art is unable to cope with collectively: i) existing metrics are not comprehensive;
ii) XAI techniques are highly heterogeneous; iii) misinterpretations are normally rare events. 
To tackle these challenges, we introduce two black-box evaluation methods, concerning the \textit{worst-case} interpretation discrepancy and a \textit{probabilistic} notion of \textit{how robust in general}, respectively.
Genetic Algorithm (GA) with bespoke fitness function is used to solve constrained optimisation for efficient \textit{worst-case} evaluation. 
Subset Simulation (SS), dedicated to estimate rare event probabilities, is used for evaluating \textit{overall} robustness. Experiments show that the accuracy, sensitivity, and efficiency of our methods outperform the state-of-the-arts. Finally, we demonstrate two applications of our methods: ranking robust XAI methods and selecting training schemes to improve both classification and interpretation robustness.

\end{abstract}
\section{Introduction}

\begin{figure}[!htbp]
	\centering
	\includegraphics[width=\linewidth]{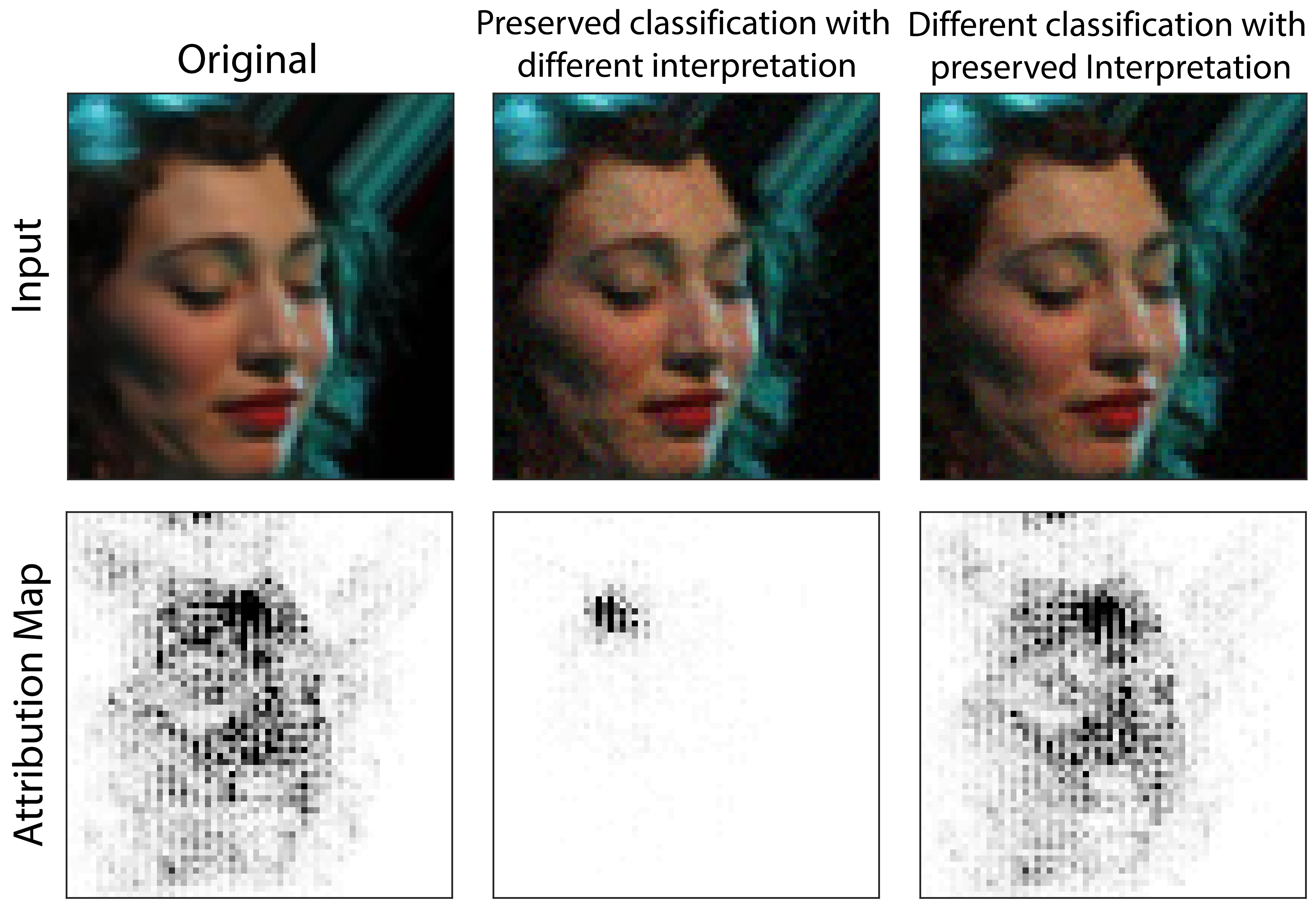}
	\caption{Two types of misinterpretations after perturbation}
	\label{fig:illustration}
\end{figure}

A key impediment to the wide adoption of Deep Learning (DL) is its perceived lack of transparency. Explainable AI (XAI) is a research area that aims at providing the visibility into how a DL model makes decisions, and thus enables the use of DL in vision-based safety critical applications, such as autonomous driving \cite{omeiza2021explanations}, and medical image analysis \cite{van2022explainable}. Typically, XAI techniques visualise which input features are significant to the DL model's prediction via attribution maps \cite{arrieta_explainable_2020,huang_survey_2020}. However, interpretations\footnote{Despite the subtle difference between interpretability and explainability, we use both terms interchangeably as attributes of DL models in this paper. However, as suggested in \cite{molnar_interpretable_2020}, we use the terms explanation/interpretation specifically for individual predictions.} suffer from the lack of robustness. Many works have shown that a small perturbation can manipulate the interpretation while keeping model's prediction unchanged, e.g., \cite{ghorbani2019interpretation,kindermans_reliability_2019}. Moreover, there exists the misinterpretation of Adversarial Examples (AEs) \cite{zhang_interpretable_2020}, i.e., adversarial inputs are misclassified\footnote{Without loss of generality, in this paper we assume the DL model is a classifier if with no further clarification.} by the DL model, but interpreted highly similarly to the benign counterparts. 
Fig.~\ref{fig:illustration} illustrates examples of the aforementioned two types of misinterpretations.
In this regard, it is vital to assess how robust the coupled DL model and XAI method are against input perturbations, which motivates this work. 

To answer the question, the first challenge we recognise is the lack of diverse evaluation metrics from the state-of-the-art.
Most of the existing works focus on adversarial attack \cite{heo2019fooling} and defence \cite{dombrowski2022towards,tang2022defense} on explanations, which essentially answer the \textit{binary} question of whether there exist any adversarial interpretation in some perturbation distance. On the other hand, evaluation methods mainly study \textit{worst-case} metrics, e.g., the \textit{maximum} change in the resulting explanations when perturbations are made \cite{alvarez2018robustness} and local \textit{Lipschitz continuity} as the sensitivity to perturbations \cite{yeh2019fidelity}.  However, for \textit{systematic} evaluation, we also need a notion of \textit{how robust in general} the model is whenever a misinterpretation can be found (in line with the insight gained from evaluating classification robustness \cite{webb2018statistical}). We introduce two metrics concerning the \textit{worst-case} interpretation discrepancy and a probabilistic metric to calculate the \textit{proportion} of misinterpretations in the local norm-ball around the original input, that complement each other from different perspectives.

Second, XAI techniques are so heterogeneous that no existing white-box evaluation methods are generic enough to be applicable to all common ones. That said, black-box methods, that only access inputs and outputs of the coupled DL model and XAI tool without requiring any internal information, are promising for all kinds of XAI techniques (including perturbation-based ones that are missing from current literature). Based on this insight, we design a Genetic Algorithm (GA) and a statistical Subset Simulation (SS) approach to estimate the aforementioned two robustness metrics, both of which are of black-box nature. 


The third challenge we identified is that misinterpretations are normally rare events in a local norm-ball. Without white-box information like gradients, black-box methods have to leverage auxiliary information to detect such rare events efficiently. To this  end, we design bespoke fitness functions in the GA (when solving the optimisation) and retrofit the established SS (dedicated to estimating rare event probabilities \cite{au2001estimation}) for efficient evaluation.

To the best of our knowledge, no state-of-the-art methods can collectively cope with the three pinpointed challenges like ours. To validate this claim, we conduct experiments to study the accuracy, sensitivity, and efficiency of our methods. Additionally, we develop two practical applications of our methods: i) We evaluate a wide range of XAI techniques and gain insights that no XAI technique is superior in terms of robustness to both types of adversarial attacks. ii) We discover a strong correlation between classification robustness and interpretation robustness through theoretical analysis (see Appx.) and empirical studies. We also identified the best training scheme to improve both aspects.

In summary, the key contributions of this paper include:
\begin{itemize}
	\item Two diverse metrics, worst-case interpretation discrepancy and probabilistic interpretation robustness, complement each other as a versatile approach, allowing for a holistic evaluation of interpretation robustness. 
	\item We introduce new methods based on GA and SS to estimate these two metrics. These methods are black-box and thus applicable to diverse XAI tools, enabling robustness evaluation of perturbation-based XAI techniques for the first time. Despite the rare occurrence of misinterpretations, our GA and SS algorithms efficiently detect them.
	\item We demonstrate two practical applications of our methods: ranking robust XAI techniques and selecting training schemes to improve both classification and interpretation robustness.
\end{itemize}







\section{Related Work}
\textbf{Evaluation of Interpretation Robustness:}
Existing evaluation metrics, proposed for interpretation robustness, only consider the misinterpretation when the prediction label of perturbed inputs remains unchanged \cite{alvarez2018robustness}. \cite{alvarez2018robustness} estimates the Local Lipschitz of interpretation, while \cite{yeh2019fidelity} introduces the max-sensitivity and average-sensitivity of interpretation. Both of them use Simple Monte Carlo (SMC) sampling to estimate their metrics. \cite{wicker2022robust} formally certify the robustness of gradient-based explanation by propagating a compact input or parameter set as symbolic intervals through the forwards and backwards computations of the neural network (NN). In \cite{dasgupta2022framework}, it defines the consistency as the probability that the inputs with the same interpretation have the same prediction label. However, their evaluation method is only applicable to tree ensemble models and tabular datasets, leaving the probabilistic estimation of misinterpretation for high-dimensional image datasets blank. Notably, toolsets/benchmarks \cite{kokhlikyan2020captum,agarwal2022openxai} for evaluating XAI techniques are emerging in the last two years. They are not specifically built for evaluating interpretation robustness, thus only concern the aforementioned metrics. That said, our metrics and their efficient estimators can be integrated into and complement those toolsets/benchmarks.

\textbf{Adversarial Attack and Defence on Interpretation:}
Ghorbani et al. first introduce the notion of adversarial perturbation to NN interpretation \cite{ghorbani2019interpretation}. Afterwards, several works are dedicated to generating indistinguishable inputs which have the same prediction label but substantially different interpretations \cite{heo2019fooling,slack2020fooling}. The theoretical analysis has shown that the lack of interpretation robustness is related to geometrical properties of NNs \cite{dombrowski2019explanations}. In \cite{zhang_interpretable_2020}, a new class of attack is proposed to fool the NN's prediction as well as the coupled XAI method. GA is introduced to manipulate SHAP in \cite{baniecki2022manipulating}.
In \cite{dombrowski2019explanations}, an upper bound on maximum changes of gradient-based interpretation is derived. The upper bound is proportional to the smooth parameter of the softplus activation function, which can be smoothed to improve the interpretation robustness. In \cite{dombrowski2022towards}, regularisation on training, like weight decay, and minimising the hessian of NNs are theoretically proved to be effective for training more robust NNs against interpretation manipulation. In \cite{zhao_baylime_2021}, prior knowledge, e.g., from V\&V evidence, is used in Bayesian surrogate models for more robust and consistent explanations. Specifically designed for perturbation-based XAI tools, \cite{carmichael2022unfooling} devises defenses against adversarial attack.

\section{Preliminaries}
\subsection{Feature-Attribution based XAI}
\label{sec_input_interpretation}
While readers are referred to \cite{arrieta_explainable_2020} for a review, we list common feature-attribution based XAI methods \cite{zhang2021survey,huang_survey_2020} that are studied by this work.
For gradient-based methods, we consider the Guided Backpropagation, Gradient~$\times$~Input, Integrated Gradients, GradCAM, LRP and DeepLift. For perturbation-based methods, we study LIME and SHAP. Descriptions with greater detail of these XAI methods are presented in Appx.~\ref{appendix_input_interpretation}.



\subsection{Local Robustness of Interpretation}
\label{sec:robust_interpretation}

Analogous to the adversarial robustness of classification, interpretation can be fooled by adding perturbations to the input. The interpretation robustness is highly related to the robustness of classification, since the attribution map is produced based on some prediction class. Therefore, we first define the robustness of classification and then formalise the robustness of interpretation,
using the following notations. Given an input seed $x$, we may find a norm ball $B(x,r)$ with the central point at $x$ and radius $r$ in $L_p$ norm. 
We denote the prediction output of the DL model as the vector $f(x)$ with size equal to the total number of labels.

Classification robustness requires that DL model's prediction output should be invariant to the human imperceptible noise, which can be expressed through the prediction loss around an input seed $x$ 
\begin{equation}
\label{eq_pred_loss}
\begin{split}
    J(f(x),f(x')) = \max_{i \neq y} (f_i(x') - f_y(x')) \\
    y = \argmax_i f_i(x), \quad x' \in B(x,r)
\end{split}
\end{equation}
where $f_i(x')$ returns the probability of label $i$ after input $x'$ being processed by the DL model $f$. Note, $J \geq 0$ implies that $x'$ is an AE. We then define the following indicator function for misclassification within the norm ball $B(x,r)$ 
\begin{equation}
\label{indicator_classification}
    I_c = 
    \begin{cases}
      -1 & \text{if $J(f(x),f(x')) \geq 0$}\\
      1 & \text{if $J(f(x),f(x')) < 0$}\\
    \end{cases} 
\end{equation}
That is, $I_c=-1$ indicates misclassification, otherwise $1$.

Previous works study two circumstances when small perturbation fools the interpretation $g(x)$, cf. Fig.~\ref{fig:illustration} for examples.
We use the interpretation discrepancy $\mathfrak{D}(g(x),g(x'))$ (defined later) to quantify the difference between the new interpretation $g(x')$ after perturbation and the reference $g(x)$, where $x' \in B(x,r)$. We then introduce two constants as thresholds, $\alpha$ and $\beta$, such that $\mathfrak{D} < \alpha$ represents consistent interpretations, 
while $\mathfrak{D} > \beta$ represents inconsistent interpretations\footnote{When $\alpha \leq \mathfrak{D} \leq \beta$, it represents the case that we cannot clearly decide if the two interpretations are consistent or not.}. 
Two misinterpretation regions within the norm ball $B(x,r)$ are then defined as
\begin{equation}
\label{eq:misinterpretation}
\widehat{F} = \{\mathfrak{D} >\beta \land J < 0\}, \quad \widetilde{F} = \{\mathfrak{D} < \alpha \land J \geq0\} 
\end{equation}


$\widehat{F}$ represents preserved classification with different interpretation and $\widetilde{F}$ represents different classification with preserved interpretation, respectively. Note, $\alpha$ and $\beta$ are hyper-parameters that define the consistency notion of interpretations. They may vary case by case in the specific application context, representing the level of strictness required by the users on interpretation robustness. For example, if we use PCC (defined later) to quantify $\mathfrak{D}$, i.e. $\mathfrak{D}$=1/PCC, there is a rule of thumb \cite{akoglu2018user} that $\text{PCC} < 0.4$ ($\beta=1/0.4$) indicates inconsistent interpretations while $\text{PCC}>0.6$ ($\alpha = 1/0.6$) represents consistent interpretations.



\subsection{Interpretation Discrepancy Metrics}
In order to quantify the visual discrepancy between the XAI results (i.e., attribution maps), there are several commonly used metrics, including Mean Square Error (MSE), Pearson Correlation Coefficient (PCC), and Structural Similarity Index Measure (SSIM) \cite{dombrowski2019explanations}. PCC and SSIM have the absolute values in $[0, 1]$. The smaller values indicate the larger discrepancy between two interpretations. MSE calculates the average squared differences, the value of which more close to 0 means higher similarity. Then, interpretation discrepancy $\mathfrak{D}$ can be expressed as
\begin{equation}
    \mathfrak{D} = \frac{1}{\text{PCC}} \: \text{ or } \: \frac{1}{\text{SSIM}} \: \text{ or } \: \text{MSE}
\end{equation}

\section{Worst Case Evaluation}


The conventional way to evaluate robustness of classification is based on the \textit{worst case} loss under the perturbation \cite{yu2019interpreting}. This underlines the adversarial attack and motivates the adversarial training. Similarly, \textit{the worst case interpretation discrepancy between the original input and perturbed input may reflect the interpretation robustness.}

There are two types of misinterpretations after perturbation in a local region, cf. Eq.~\eqref{eq:misinterpretation}. Accordingly, two optimisations are formalised for the worst case interpretation discrepancy:
\begin{equation}
\label{first_int}
\begin{split}
   sol_{\widehat{F}} =  &\max_{x' \in B(x,r)} \; \mathfrak{D}(g(x),g(x')) \\
    &s.t. \; J(f(x),f(x')) < 0
\end{split}
\end{equation}
\begin{equation}
\label{second_int}
\begin{split}
    sol_{\widetilde{F}} = &\min_{x' \in B(x,r)} \; \mathfrak{D}(g(x),g(x')) \\
    &s.t. \; J(f(x),f(x')) \geq 0
\end{split}
\end{equation}
That is, $sol_{\widehat{F}}$ corresponds to finding the largest interpretation discrepancy when perturbed input is still correctly classified. While $sol_{\widetilde{F}}$ is the minimum interpretation discrepancy between the AE $x'$ and input seed $x$.



Previous works adopt white-box methods to solve the above optimisations for adversarial explanations 
\cite{zhang_interpretable_2020,ghorbani2019interpretation}, in which case the DL model $f(x)$ and XAI method $g(x)$ are required to be fully accessible to their internal information. In addition, many XAI methods $g(x)$ are non-differentiable, and the strong assumptions (like smoothing gradient of ReLU non-linearity) are made to enable derivative-based optimisation. In contrast, Genetic Algorithm (GA) is a derivative-free method for solving both constrained and unconstrained optimisations, and has been successfully applied to the evaluation of classification robustness \cite{chen2019poba}. That motivates us to develop a black-box evaluation method for interpretation robustness based on GA. GA consists of 5 steps: \textit{initialisation}, \textit{selection}, \textit{crossover}, \textit{mutation}, and \textit{termination}, the middle three of which are repeated until the convergence of fitness function values. We refer readers to Appx.~\ref{appendix_GA} for more details of GA.

\textbf{Initialisation:} The population with $N$ samples is initialized. Diversity of initial population could promise approximate global optimal \cite{konak2006multi}. Normally, we use the Gaussian distribution with the mean at input seed $x$, or a uniform distribution to generate a set of diverse perturbed inputs within the norm ball $B(x,r)$.

\textbf{Selection:} The core of GA is the design of fitness functions. Fitness function guides the selection of parents for latter operations. Considering the constrained optimization, we design the fitness function based on the superiority of feasible individuals to make distinction between feasible and infeasible solutions \cite{10.5555/645513.657601}. For the optimisation of Eq.~\eqref{first_int}, the constraint can be directly encoded as the indicator $I_c$ into the fitness function
\begin{equation}
    \mathcal{F} (x') =  I_c \, \mathfrak{D}(g(x),g(x'))
\end{equation}
and $\mathfrak{D}(g(x),g(x'))$ is always none negative. All feasible individuals satisfying the constraint $J(f(x),f(x')) < 0$ will have $I_c = 1$, and $\mathcal{F} > 0$. If the constraint is violated, then $I_c = -1$, and $\mathcal{F} < 0$. In other words, the individuals violating the constraint will have smaller fitness values than the others and are suppressed during the evolution.

For the optimisation of Eq.~\eqref{second_int}, we note $J > 0$ is a rare event within the local region $B(x,r)$, as AEs are normally rare \cite{webb2018statistical}. To accelerate the search in the feasible input space, we set two fitness functions $\mathcal{F}_1$ and $\mathcal{F}_2$. 
$\mathcal{F}_1$ increases the proportion of AEs in the population. On this basis, when over half amount of the population are AEs, 
$\mathcal{F}_2$ will guide the generation of adversarial explanations.
\begin{equation}
    \mathcal{F}_1 (x') = J(f(x),f(x')) \quad \mathcal{F}_2 (x') = -I_c/\mathfrak{D}(g(x),g(x')) 
\end{equation}
In $\mathcal{F}_2$, $I_c$ also penalises the violation of constraints, which keeps the optimisation conditioned on AEs. Instead of directly selecting the best fitted individuals, we choose the fitness proportionate selection \cite{lipowski2012roulette}, which can maintain good diversity of population and avoid premature convergence. Then, the probability of selection $p_i$ for each individual $x'_i$ is formulated as
\begin{equation}
    p_i = \frac{\mathcal{F}(x'_i)}{\sum_{j=1}^{N} \mathcal{F}(x'_j)} 
\end{equation}

\textbf{Crossover:} The crossover operator will combine a pair of parents from last step to generate a pair of children, which share many of the characteristics from the parents. The half elements of parents are randomly exchanged.

\textbf{Mutation:} Some elements of children are randomly altered to add variance in the evolution. It should be noticed that the mutated samples should still fall into the norm ball $B(x,r)$. Finally, the children and parents will be the individuals for the next iteration.

\textbf{Termination:} GA terminates either when the allocated computation budget (maximum number of iterations) is depleted or the plateau is reached such that successive iterations no longer produce better results. 

\section{Probabilistic Evaluation}

\subsection{Probabilistic Metrics}
In addition to the worst case evaluation, probabilistic evaluation based on statistical approaches is of the same practical interest---a lesson learnt from evaluating classification robustness \cite{webb2018statistical,wangUAI21oxford} and DL reliability \cite{zhao_assessing_2021,dong_reliability_2022}. Thus, we study the \textit{probability of misinterpretation} within $B(x,r)$, regarding the two types of misinterpretations\footnote{Through out the paper, we use the shorthand notation $F$ for either $ \widehat{F}$ or $\widetilde{F}$, according to the context.} of the input image $x$ under study:
\begin{equation}
\label{eq:probability_eval}
P_F(x)= \int_{x' \in B(x,r)} \mathbbm{1}_{x' \in F} \, q(x') \, dx', \quad F = \widehat{F} \, \text{ or } \, \widetilde{F}
\end{equation}
where $x'$ is a perturbed sample under the local distribution $q(x')$ (precisely the ``input model'' used by \cite{webb2018statistical}, when studying local probabilistic metric) in $B(x,r)$. $\mathbbm{1}_{x' \in F}$ is equal to $1$ when $x' \in F$ is true, $0$ otherwise. Intuitively, Eq.~\eqref{eq:probability_eval} says, for the given input image $x$, if we generate an infinite set of perturbed samples locally (i.e., within a norm ball $B(x,r)$) according to the distribution $q$, then the proportion of those samples fall into the misinterpretation region $F$ is defined as the proposed probabilistic metric.

\subsection{Estimation by Subset Simulation}
To estimate the two probabilistic metrics defined by Eq.~\eqref{eq:probability_eval}, there are two challenges: i) misinterpretations represented by $\widetilde{F}$ and $\widehat{F}$ are arguably rare events (that confirmed empirically later in our experiments);
ii) inputs of DL models are usually high dimensional data, like images. The first challenge requires sampling methods specifically designed for rare events rather than SMC (that is known to be inefficient for rare events). The second challenge rules out some commonly used advanced sampling methods, like importance sampling, as they may not be applicable to high dimensional data \cite{au2003important}.

The well-established Subset Simulation (SS) can efficiently calculate the small failure probability in high dimensional space \cite{au2001estimation} and has been successfully applied to assessing classification robustness of DL models \cite{webb2018statistical}. As a black-box method, it only involves the input and response of interest for calculation, thus generic to diverse XAI methods $g(x)$. 
The main idea of SS is introducing intermediate failure events so that the failure probability can be expressed as the product of larger conditional probabilities. Let $F=F_m \subset F_{m-1} \subset \cdots \subset F_2 \subset F_1$ be a sequence of increasing events so that $F_m = \bigcap_{i=1}^{m}F_i$. By conditional probability, we get
\begin{equation}
\label{eq:subset_simulation}
    P_F \! :=\! P(F_m)\! =\! P(\bigcap_{i=1}^{m} \! F_i) 
        \! = \! P(F_1)\!\prod_{i=2}^{m}P(F_{i}|F_{i-1})
\end{equation}
The conditional probabilities of intermediate events involved in Eq.~\eqref{eq:subset_simulation} can be chosen sufficiently large so that they can be efficiently estimated. For example, $P(F_1)=1$, $P(F_i|F_{i-1})=0.1$, $i=2,3,4,5,6$, then $P_F \approx 10^{-5}$ which is too small for efficient estimation by SMC sampling. In this section, we adapt SS for our problem as what follows.

\subsubsection{Design of Intermediate Events}
$\widehat{F}$ and $\widetilde{F}$ can be decomposed as the series of intermediate events through the expression of property functions $J$ and $\mathfrak{D}$. For $\widehat{F}$, $J <0$ is not rare for a well-trained DL model, representing the correctly classified input after perturbation. Thus, the intermediate events $\widehat{F}_{i-1}$ and $\widehat{F}_{i}$ can be chosen as
\begin{equation}
\begin{split}
    \widehat{F}_{i-1} = &\{ I_c\mathfrak{D}  > \beta_{i-1}\}, \quad \widehat{F}_{i} = \{I_c\mathfrak{D}>\beta_{i}\} \\
    & \text{where} \quad  \beta_{i-1} < \beta_{i} \leq \beta
\end{split}
\end{equation}
such that $\widehat{F}_{i} \subset \widehat{F}_{i-1}$. $I_c$ (in Eq.~\ref{indicator_classification}) encodes the constraint $J <0$ as the sign of $\mathfrak{D}$. 

In contrast, $J \geq 0 $ in $\widetilde{F}$ represents the occurrence of AEs that are rare events, which cannot be directly expressed as the indicator $I_c$,
since the random sampling within $B(x,r)$ cannot easily satisfy $J \geq 0 $. Thus, for $\widetilde{F}$, $J \geq 0 $ should be chosen as the critical intermediate event.
\begin{equation}
    \widetilde{F}_j = \{J \geq 0\}, \quad \text{where} \quad 1<j<m
\end{equation}
For intermediate events $\widetilde{F}_{i-1}$ and $\widetilde{F}_{i}$, when $i \! < j$, we set 
\begin{equation}
\begin{split}
    \widetilde{F}_{i-1} & = \{J > \gamma_{i-1}\} , \quad \widetilde{F}_{i} = \{J > \gamma_{i}\} \\
    & \text{where} \quad \gamma_{i-1} < \gamma_{i} < 0
\end{split}
\end{equation}
such that $\widetilde{F}_{j} \subset \widetilde{F}_{i} \subset \widetilde{F}_{i-1}$. And for intermediate events $\widetilde{F}_{k-1}$ and $\widetilde{F}_{k}$, when $k-1 > j$, we can set 
\begin{align}
    \widetilde{F}_{k-1}  = \{-I_c /\mathfrak{D} & > 1/\alpha_{k-1}\}, \quad  \widetilde{F}_{k}  = \{-I_c/ \mathfrak{D} > 1/\alpha_{k} \} \nonumber \\
    &\text{where} \quad 0 < \alpha \leq \alpha_{k} < \alpha_{k-1}
\end{align}
such that $\widetilde{F}_{k} \subset \widetilde{F}_{k-1} \subset \widetilde{F}_{j}$.

\subsubsection{Estimating Conditional Probabilities}
Upon formally defined intermediate events, the question arises on how to set $\beta_{i}$, $\gamma_{i}$ and $\alpha_{i}$ to make the conditional probability $P(F_{i}|F_{i-1})$ sufficiently large for estimation by a few simulations. Also, simulating new samples from $F_i$ for estimating next conditional probability $P(F_{i+1}|F_{i})$ is difficult due to the rarity of $F_{i}$. Therefore, the Markov Chain Monte Carlo sampling based on the Metropolis–Hastings (MH) algorithm is adopted. For simplicity, the intermediate event threshold is generally denoted as $L_i = \{\beta_i, \gamma_i, \alpha_i\}$.

\subsubsection{Choices of Intermediate Event Threshold}
Start from estimating $P(F_1)$, $F_1$ is chosen as the common event such that $N$ samples are drawn from $q(\cdot)$ by SMC and all belong to $F_1$. A feasible way is setting the threshold of property function $L_1$ to $-\infty$, and $P(F_1) = 1$. For $i = 2, \cdots, m$, $L_i$ affects the values of condition probabilities and hence the efficiency of SS. It is suggested that  $L_i$ is set adaptively to make $P(F_{i}|F_{i-1})$ approximately equals to $\rho$, and $\rho$ is a hyper-parameter in SS (that takes a decimal less than 1 and normally $\rho=0.1$ yields good efficiency, although it can be empirically optimised),
i.e., $ P(F_{i}|F_{i-1}) \approx \rho $.
That is, at each iteration $i-1$ when we simulate $N$ samples, $\rho N$ samples should belong to $F_{i}$.

\subsubsection{Simulating New Samples from $q(\cdot|F_i)$} 

At iteration $i =2, \cdots, m-1$, we already have $\rho N$ samples belonging to $F_i$ and aim to simulate new samples to enlarge the set to $N$, so that the next conditional probability $P(F_{i+1}|F_i) = \frac{1}{N} \sum_{k=1}^N \mathbbm{1}_{F_{i+1}}(x'_k)$ can be calculated. We can pick up an existing sample $x'$ subject to the conditional distribution $q(\cdot|F_i)$,
denoted as $x' \sim q(\cdot|F_i)$, and use the Metropolis
Hastings (MH) algorithm to construct a Markov Chain. By running $M$ steps of MH, the stationary distribution of the Markov Chain is $q(\cdot|F_i)$. Then new data $x'' \sim q(\cdot|F_i)$ can be sampled from the Markov Chain and added into the set. More details of the MH algorithm for SS are presented in Appx.~\ref{appendix_ss}. 

\subsubsection{Termination Condition and Returned Estimation}

After the aforementioned steps, SS divides the problem of estimating a rare event probability into several simpler ones---a sequence of intermediate conditional probabilities as formulated in Eq.~\eqref{eq:subset_simulation}. The returned estimation $\widebar{P}_F$ and coefficient of variation (c.o.v.) $\widebar{\delta}$ (measuring the estimation error) are
\begin{equation}
\label{mean_cov}
    \widebar{P}_F = \prod_{i=1}^{m} \frac{1}{N} \sum_{k=1}^N \mathbbm{1}_{F_i}(x'_k), \quad
    \widebar{\delta}^2 \approx \sum_{i=1}^m \frac{1-\widebar{P}_{F_i}}{\widebar{P}_{F_i} N}(1+\lambda_i)
\end{equation}
where $\lambda_i >0$ represents the efficiency of the estimator using dependent samples drawn from the Markov Chain. For simplicity, we can assume $\lambda_i \approx 0$ when the number of steps $M$ of MH is large
\cite{cerou2012sequential}. Since each conditional probability $P(F_{i}|F_{i-1})$ approximately equals to $\rho$, then by Eq.~\eqref{eq:subset_simulation}, the returned estimation $\widebar{P}_F \approx \rho^{m-1}$. $m$ is the total number of intermediate event generated adaptively. The adaptive generation of intermediate events terminates when $\widebar{P}_F<P_{\text{min}}$, and $P_{\text{min}}$ is a given termination threshold. More details of statistical properties of the estimator, like error bound, efficiency are presented in Appx.~\ref{appendix_ss}.


%

\section{Experiments}







\subsection{Experiment Setup}
We consider three public benchmark datasets
, five XAI methods, and five training schemes in our experiments. The norm ball radius, deciding the oracle of robustness, is calculated with respect to the $r$ separation property \cite{yang2020closer}. That is, $r = 0.3$ for MNIST, $r = 0.03$ for CIFAR10, and $r = 0.05$ for CelebA. More details of the DL models under study are presented in Appx.~\ref{exp_model_details}. For the probabilistic evaluation using SS, without loss of generality, we consider the uniform distribution as $q(x')$ within each norm ball. We compare $\mathfrak{D} = $ MSE, 1/PCC, and 1/SSIM for measuring interpretation discrepancy in Appx.~\ref{appendix_interpretation_discrepancy}, and find PCC is better to quantify the interpretation difference in our cases. Based on sensitivity analysis, we choose hyperparameters PCC thresholds $1/\beta  = 0.4$, $1/\alpha = 0.6$, MH steps $M = 250$, $\rho=0.1$, $\ln P_{\text{min}} = -100$ for probabilistic evaluation, and population size $N = 1000$, number of iteration $itr = 500$ for the worst case evaluation by GA. Our tools and experiments are publicly available at \url{https://github.com/havelhuang/Eval\_XAI\_Robustness}.



\subsection{Sensitivity to Hyper-Parameter Settings}

\begin{figure}[!ht]
\centering
\includegraphics[width=\linewidth]{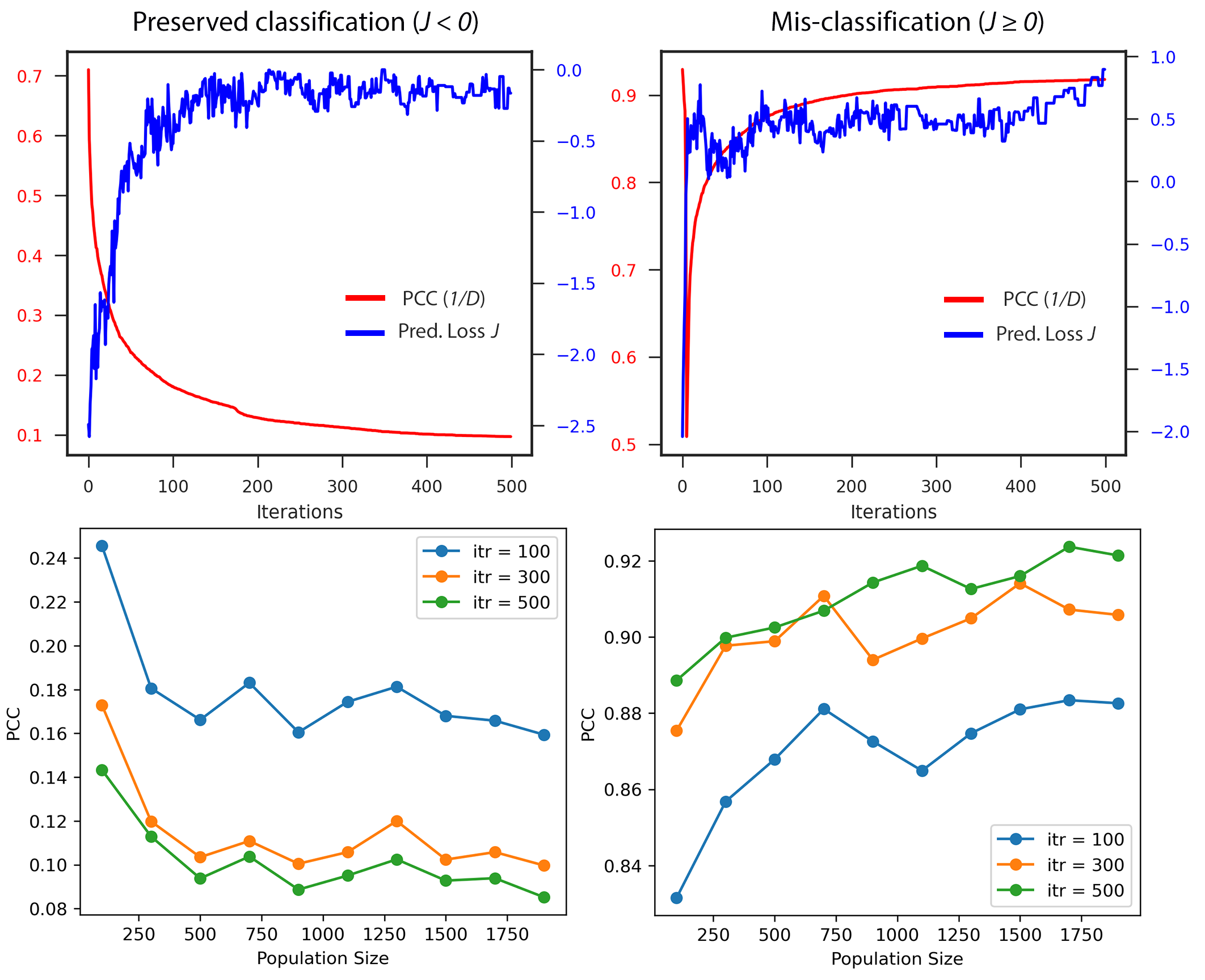}
\caption{Sensitivity of objective $\mathfrak{D}$ and constraint $J$ to GA's population size and iteration numbers. Each column represents a type of misinterpretation. 
1st row: quickly converged GA objectives satisfying the constraint, with fixed population size of 1000 and varying iterations. 2nd row: GA solutions, with fixed iteration numbers and varying population size. A test seed (representing a norm ball) from CelebA is used; interpretation discrepancy $\mathfrak{D}$ is measured by $1/$PCC; ``Gradient$\times$Input'' XAI method is studied.}
\label{fig:ga_accuracy}
\end{figure}

We first investigate the sensitivity of objective function $\mathfrak{D}$ and constraint $J$ (cf. Eq.~\eqref{first_int} and \eqref{second_int}) to GA's population
size and iteration numbers, as shown in Fig.~\ref{fig:ga_accuracy}. We observe from the 1st row that interpretation discrepancy measured by PCC (the red curve) quickly converge after 300 iterations with the satisfaction of constraint $J$ (the blue curve), showing the effectiveness of our GA. From the 2nd row, we notice that the optimisation is not sensitive to population size, compared with the number of iterations, i.e., population size over 500 cannot make significant improvement to the optimisation. In addition, if the number of iterations is sufficiently large, 
the effect of population size on optimal solution is further diminished. We only present the results of one seed from CelebA, cf.~Appx.~\ref{appendix_parameter} for more seeds from other datasets, while the general observation remains.


\begin{figure}[!htbp]
\centering
\includegraphics[width=\linewidth]{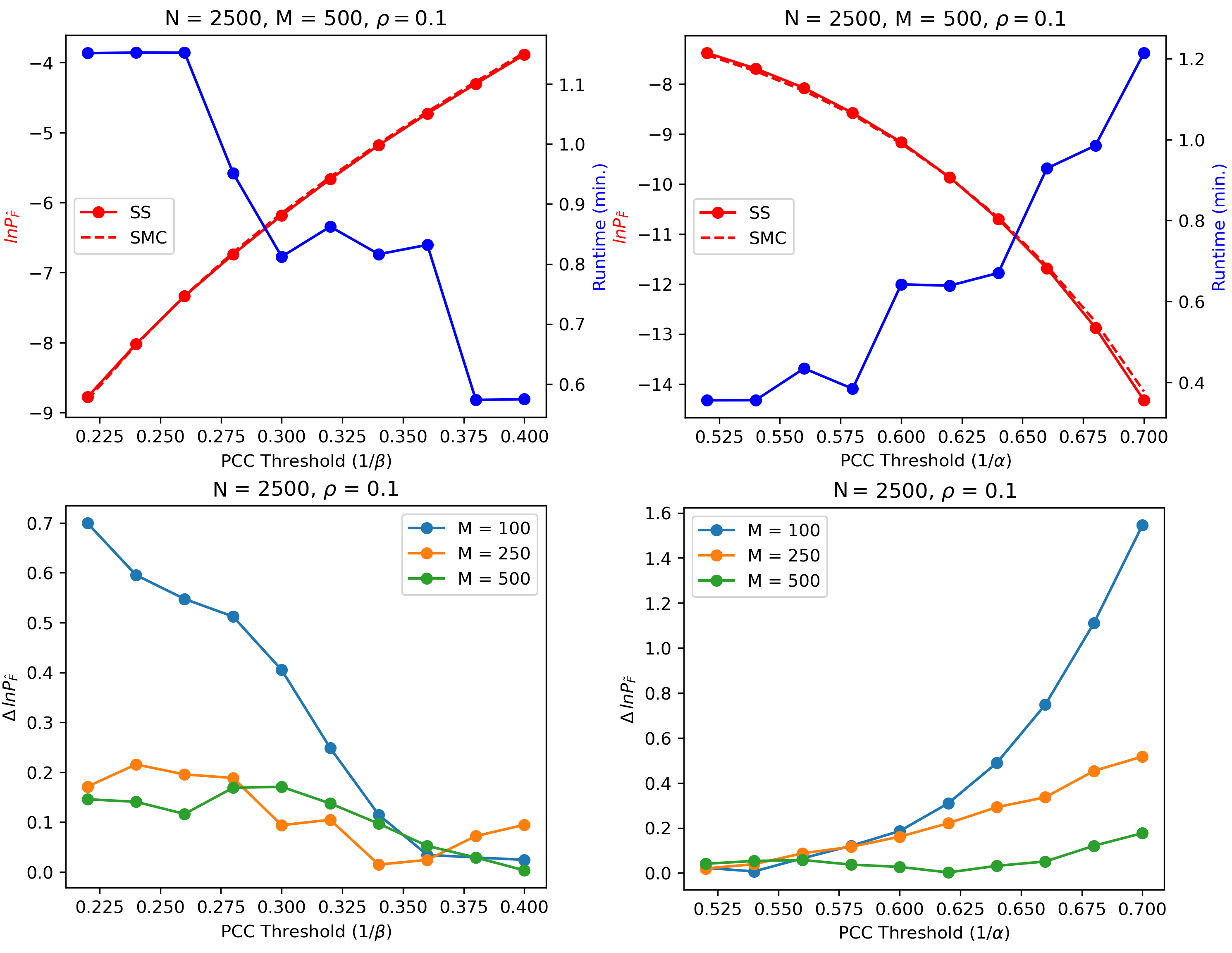}
\caption{
Each column represents a type of misinterpretation.
1st row: the probability of misinterpretation ($\ln P_F$) estimations returned by
SS and approximated ground truth by SMC\footref{ftnote_sample_smc}, 
varying the rarity of misinterpretations.
Overlapping of two red curves shows high accuracy of SS.
2nd row: sensitivity of SS accuracy $\Delta \ln P_F$ to MH steps $M$, varying the rarity level of misinterpretation controlled by PCC threshold. A test seed from MNIST and ``Gradient$\times$Input'' XAI method  are used; Results are averaged over 10 runs.}
\label{fig:ss_accuracy}
\end{figure}

Next, we study the sensitivity of SS accuracy to the number of MH steps $M$, varying the PCC threshold that defines the rarity level of misinterpretation events.
In Fig.~\ref{fig:ss_accuracy}, we can calculate the difference $\Delta \ln P_F$ between SS estimations and the approximated ground truth (by SMC estimations using a sufficiently large number of samples\footnote{\label{ftnote_sample_smc}We use $10^8$ samples (for the specific seed) which can accurately estimate a small probability in natural logarithm around $-17\!\sim\!-18$.}). The 1st row shows the overlapping of SS and SMC estimations (two red curves) and the reducing running time (the blue curve) when decreasing the rarity levels of misinterpretations (by controlling the PCC threshold). From the 2nd row we observe that, with increased MH steps $M$, the estimation accuracy of SS is significantly improved. In addition, the rarity of misinterpretation events determines the choice of $M$. E.g., if $\ln P_{\widehat{F}} = -3.87$ with $\widehat{F} = \{\textit{PCC} < 0.4 \land J <0 \}$, then $M = 100$ already achieves high precision without additional sampling budget. Other parameters, e.g. the number of samples $n$ and sample quantile $\rho$ that are discussed in Appx.~\ref{appendix_parameter}, are in general less sensitive than the number of MH steps $M$. 

In summary, sensitivity analysis provides the basis of setting hyper-parameters in later experiments: 500 iterations and 1000 population size for GA, 250 MH steps for SS.


\subsection{Accuracy and Efficiency of Evaluation}

We study the accuracy of our GA-based evaluation, comparing with state-of-the-art \cite{alvarez2018robustness,yeh2019fidelity}---they define the local Lipschitz ($\textit{SENS}_{\textit{LIPS}}$) and max-sensitivity ($\textit{SENS}_{\textit{MAX}}$) metrics for the maximum interpretation discrepancy, and empirically estimate the metrics using SMC sampling. For fair comparisons, we first choose MSE as the interpretation discrepancy metric in our fitness functions of GA, and then apply both GA and SMC to generate two populations of interpretations in which we calculate the three robustness metrics respectively and summarise in Table~\ref{tab:metrics_comparison}. We use $5 \! \times \! 10^5$ samples for both GA and SMC.



\begin{table}[h]
\centering
\caption{Three worst case robustness metrics estimated by our GA and SMC, averaged over 100 test seeds. GA outperforms SMC (used by state-of-the-arts) w.r.t. all 3 metrics.
} 
\resizebox{\linewidth}{!}{
\begin{tabular}{ccccccc}
\hline
\multirow{2}{*}{Dataset} & \multicolumn{3}{c}{GA} & \multicolumn{3}{c}{SMC} \\ \cline{2-7} 
 & \begin{tabular}[c]{@{}c@{}}MSE\\ ($sol_{\widehat{F}}$)\end{tabular} & $\textit{SENS}_{\textit{MAX}}$ & $\textit{SENS}_{\textit{LIPS}}$ & \begin{tabular}[c]{@{}c@{}}MSE 
 \end{tabular} & $\textit{SENS}_{\textit{MAX}}$ & $\textit{SENS}_{\textit{LIPS}}$ \\ \hline
MNIST & \textbf{1.549} & \textbf{36.067} & \textbf{13.747} & 0.271 & 15.226 & 2.772 \\
CIFAR10 & \textbf{42.436} & \textbf{328.147} & \textbf{314.861} & 0.589 & 38.529 & 40.232 \\
CelebA & \textbf{3.204} & \textbf{192.203} & \textbf{65.635} & 0.013 & 11.298 & 3.563 \\ \hline
\end{tabular}
}
\label{tab:metrics_comparison}
\end{table}

As shown in Table~\ref{tab:metrics_comparison}, our GA-based estimator outperforms SMC in all of the three robustness metrics. Although the metrics of local Lipschitz and max-sensitivity are not explicitly encoded as optimisation objectives in our GA, GA is still more effective and efficient to estimate those three extreme values than SMC. This is non-surprising, since all three metrics are compatible and essentially representing the same worst-case semantics. That said, our interpretation discrepancy metric complements $\textit{SENS}_{\textit{LIPS}}$ and $\textit{SENS}_{\textit{MAX}}$ (as the former is based on Lipschitz value while the latter defined only in $L_2$ norm), can be easily encoded in our GA.





In addition to the accuracy shown in Fig.~\ref{fig:ss_accuracy}, we compare the sample efficiency between SS and SMC by calculating the number of required simulations $N_{\textit{SS}}$ and $N_{\textit{SMC}}$ for achieving same estimation errors (measured by c.o.v. $\delta$). As shown in Table~\ref{tab:ss_smc_comparison}, SS requires fewer samples, showing great advantage over SMC, cf. Appx.~\ref{appendix_ss} for theoretical analysis.

\begin{table}[h]
\centering
\caption{Sample efficiency of SS and SMC. In all six cases, SS requires fewer samples ($N_{\textit{SS}}<N_{\textit{SMC}}$) than SMC for achieving the same estimation errors $\delta^2$. Each result is averaged over 10 seeds.} 
\resizebox{0.75\linewidth}{!}{
\begin{tabular}{cccccc}
\hline
Dataset & $F$ & $\ln P_{F}$ & $\delta^2$ & $N_{\textit{SS}}$ & $N_{\textit{SMC}}$ \\ \hline
\multirow{2}{*}{MNIST} & $\widehat{F}$ & -12.25 & 0.0184 & \textbf{15000} & $1.13\times10^{7}$ \\
 & $\widetilde{F}$ & -24.63 & 0.0374 & \textbf{27500} & $1.34\times10^{12}$ \\ \hline
\multirow{2}{*}{CIFAR10} & $\widehat{F}$ & -0.79 & 0.0004 & \textbf{2500} & $2500$ \\
 & $\widetilde{F}$ & -33.54 & 0.0511 & \textbf{40000} & $7.22\times10^{15}$ \\ \hline
\multirow{2}{*}{CelebA} & $\widehat{F}$ & -31.43 & 0.0482 & \textbf{35000} & $9.29\times10^{14}$ \\
 & $\widetilde{F}$ & -70.71 & 0.1090 & \textbf{80000} & $4.68\times10^{31}$ \\ \hline
\end{tabular}
}
\label{tab:ss_smc_comparison}
\end{table}




\subsection{Evaluating XAI Methods}

The first application of our methods is to draw insights on the robustness of common XAI techniques, from both the worst-case and probabilistic perspectives. Thanks to the black-box nature of GA and SS, our methods are applicable to diverse XAI tools, and we consider six popular ones in this section. In Appx.~\ref{appendix_eval_xai}, we evaluate other XAI tools and discuss how the number of perturbed samples and image segmentation affect evaluation results on LIME and SHAP (which are missing from current literature).




\begin{figure}[!htbp]
\centering
\includegraphics[width=\linewidth]{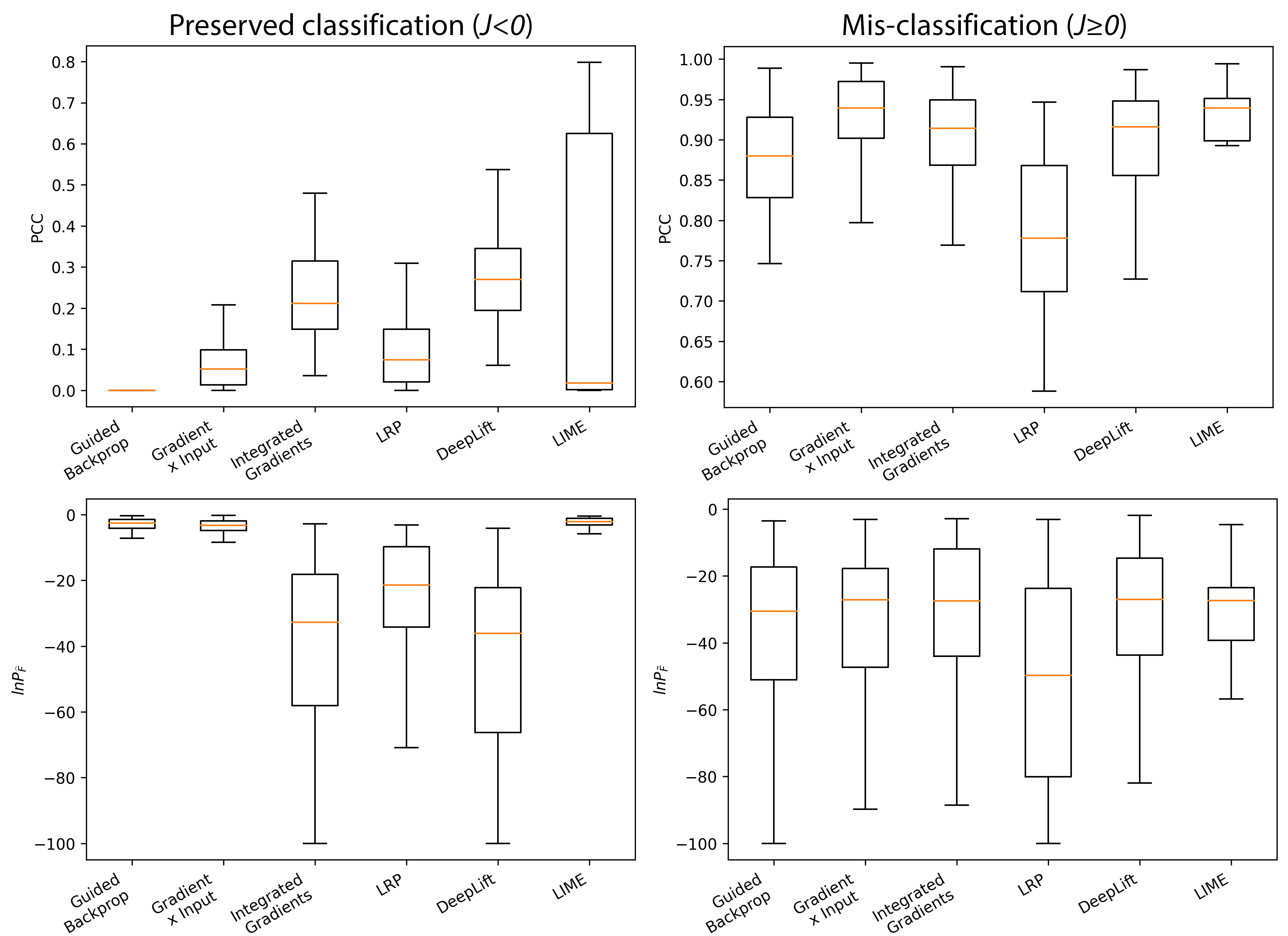}
\caption{Worst-case (1st row) and probabilistic (2nd row) robustness evaluations of five XAI methods based on 100 random seeds from MNIST. Each column represents
a type of misinterpretation---$\widehat{F}$ left and $\widetilde{F}$ right. For top-left plot, higher value means more robust; for all other plots, lower value means more robust. }
\label{fig:intepretation_eval}
\end{figure}

We randomly sample 100 seeds from MNIST for evaluations, and summarise the statistics as box-and-whisker plots in Fig.~\ref{fig:intepretation_eval}. Based on the empirical results of Fig.~\ref{fig:intepretation_eval}, we may conclude: i) Perturbation-based XAI method also suffers from the lack of robustness. ii) for misinterpretation $\widehat{F}$---correct classification ($J<0$) with inconsistent interpretation ($\textit{PCC} < 0.4$),  DeepLift and Integrated Gradients outperform others, while Guided Backprop and $\text{Gradient}\! \times\! \text{Input}$ are unrobust from both worst-case and probabilistic perspective; iii) for misinterpretation $\widetilde{F}$---wrong classification ($J \ge 0$) with persevered interpretation ($\textit{PCC} > 0.6$), while all XAI methods perform similarly w.r.t. both metrics, LRP shows better robustness than others.

The empirical insights are as expected if we consider the mechanisms behind those XAI methods. For instance, considering $\widehat{F}$, DeepLift and Integrated Gradients are more robust, since they use the reference point to avoid the discontinuous gradients (large curvature) that mislead the attribution maps \cite{shrikumar2017learning}. On the other hand, DeepLift and Integrated Gradients become vulnerable to $\widetilde{F}$. Because misclassification and misinterpretation are rare events, most perturbed inputs inside the norm ball have consistent interpretation with the seed. Consequently, the integration from the reference point which averages the attribution map over several points are prone to produce the consistent interpretations. See Appx.~\ref{appendix_eval_xai} for more discussions and experiments on CIFAR10 and CelebA dataset.

\subsection{Evaluating Training Schemes}

In this application, we study the effect of various training schemes on the interpretation robustness of DL models. In Appx.~\ref{classification_vs_interpretation}, we theoretically analyse the relation between classification robustness and interpretation robustness. The Prop.~\ref{prop_lpg} shows that input hessian norm and input gradient norm are related to the change of classification loss and interpretation discrepancy. Thus, we add input gradient and input hessian regularisation terms to the training loss, and also consider the PGD-based adversarial training that improves classification robustness through minimising the maximal prediction loss in norm balls \cite{jin2022enhancing,jin2023randomized}. Table~\ref{tab:robustness_eval} records the results.

\begin{table}[!htbp]
\centering
\caption{Evaluating classification ($c$) and interpretation ($\widehat{F}$ and $\widetilde{F}$) robustness of DL models, trained with input gradient norm regularisation (Grad. Reg.), input hessian norm regularisation (Hess. Reg.), both of them (Grad. + Hess. Reg.) and adversarial training (Adv. Train.). Results are averaged over 100 random seeds. Higher $sol_{\widehat{F}}$ means more robust, while for other metrics, the lower is the better.} 
\resizebox{0.97\linewidth}{!}{
\begin{tabular}{cc|cccccc}
\hline
\multirow{2}{*}{Dataset} & \multirow{2}{*}{Model} & \multicolumn{3}{c}{Worst Case Evaluation} & \multicolumn{3}{c}{Probabilistic Evaluation} \\ \cline{3-8} 
 &  & \begin{tabular}[c]{@{}c@{}}$sol_c$\\ (J)\end{tabular} & \begin{tabular}[c]{@{}c@{}}$sol_{\widehat{F}}$\\ (PCC)\end{tabular} & \begin{tabular}[c]{@{}c@{}}$sol_{\widetilde{F}}$\\ (PCC)\end{tabular} & $\ln P_c$ & $\ln P_{\widehat{F}}$ & $\ln P_{\widetilde{F}}$ \\ \hline
\multirow{5}{*}{MNIST} & Org. &  22.43 & 0.06 & 0.93 & -24.28 & -3.87 & -31.47  \\
 & Grad. Reg & 11.37 & 0.10 & 0.92 & -31.51 & -15.69 & -44.96 \\
 & Hess. Reg. & 10.59 & 0.17 & 0.90 & -33.36 & -21.27 & -43.85  \\
 & Grad. + Hess. & 10.04 & 0.20 & 0.90 & -36.96 & -23.79 & -46.19 \\
 & Adv. Train. &  \textbf{-0.16} & \textbf{0.21} & \textbf{0.59} & \textbf{-84.15} & \textbf{-28.67} & \textbf{-89.09}  \\ \hline
\multirow{5}{*}{CIFAR10} & Org. & 42.58 & 0.02 & 0.85 & -31.55 & -18.63 & -71.46  \\
 & Grad. Reg & 42.34 & 0.01 & 0.85 & -27.31 & -21.77 & -65.75  \\
 & Hess. Reg. & 8.99 & 0.08 & 0.81 & -76.29 & -99.20  & -91.89   \\
 & Grad. + Hess. & 8.47 & 0.06 & 0.81 & -71.65 & -98.49 & -92.39 \\
 & Adv. Train. & \textbf{-0.67} & \textbf{0.25} & \textbf{0.80} & \textbf{-92.57} & \textbf{-100} & \textbf{-95.97} \\ \hline
 \multirow{5}{*}{CelebA} & Org. & 51.08 & 0.08 & 0.86 & -13.77 & -21.58 & -70.82 \\
 & Grad. Reg & 25.29 & 0.06 & 0.88 & -45.52 & -70.22 & -83.26 \\
 & Hess. Reg. & 18.71 & 0.09 & 0.86 & -74.93 & -100 & \textbf{-95.85} \\
 & Grad. + Hess. & 25.41 & 0.06 & 0.88 & -65.95 & -100 & -94.13 \\
 & Adv. Train. & \textbf{-0.45} & \textbf{0.55} & \textbf{0.81} & \textbf{-95.09} & \textbf{-100} & -95.58 \\ \hline
\end{tabular}
}
\label{tab:robustness_eval}
\end{table}

In addition to the knowledge that input hessian can defence adversarial interpretation \cite{dombrowski2022towards}, we notice that it is 
significant and 
effective in improving both classification and interpretation robustness, than input gradient regularisation, confirming our Prop. \ref{prop_lpg}. Moreover, we discover that adversarial training is surprisingly effective at improving interpretation robustness, but at the price of dropping accuracy, cf. Appx.~\ref{exp_model_details}. This phenomenon reals the strong correlation between classification and interpretation robustness. That said, the improvement of classification robustness may lead to the improvement of interpretation robustness.

\section{Conclusion}

This paper proposes two versatile and efficient evaluation methods for DL interpretation robustness. The versatility is twofold: (1) the proposed metrics are characterising robustness from both worst-case and probabilistic perspectives; (2) GA and SS are black-box methods thus generic to heterogeneous XAI methods. Considering the rare-event nature of misinterpretations, GA and SS show high efficiency in detecting them, thanks to the bespoke design of fitness functions in GA and encoding auxiliary information as intermediate events in SS.

\paragraph{Acknowledgements}
This project has received funding from the European Union’s Horizon 2020 research and innovation programme under grant agreement No 956123.  It is also supported by the UK EPSRC through End-to-End Conceptual Guarding of Neural Architectures [EP/T026995/1], Department of Transport UK, Transport Canada and WMG center of HVM Catapult.

{\small
\bibliographystyle{ieee_fullname}
\bibliography{ref}
}

\clearpage
\newpage
\section{Appendix}
\subsection{Feature-Attribution based XAI}
\label{appendix_input_interpretation}
\paragraph{Guided Backpropagation:} It computes the gradient of output with respect to the input, but only the non-negative components of gradients are propagated to highlight the important pixels in the image \cite{springenberg2015striving}.

    
 \paragraph{Gradient $\times$ Input:} The map $g(x) = x \odot \frac{\partial f(x)}{\partial x}$ is more preferable to gradient alone to leverage the sign and strength of input to improve the interpretation sharpness \cite{shrikumar2017learning}.

    
\paragraph{Integrated Gradients:} Instead of calculating single derivative, this approach integrates the gradients from some baseline to its current input value $g(x) = (x-\bar{x}) \int_{\alpha=0}^{1} \frac{\partial f(\bar{x} + \alpha(x-\bar{x}))}{\partial x} d \alpha$, addressing the saturation and thresholding problems \cite{sundararajan2017axiomatic}.

\paragraph{GradCAM:}Gradient-weighted Class Activation Mapping (Grad-CAM) generates the visual explanation for convolutional neural network, using gradients flowing into the final convolutional layer to produce a coarse localization map, highlighting the relevant regions in the image for prediction\cite{selvaraju2017grad}.
 
 \paragraph{Layer-wise Relevance Propagation (LRP):} LRP operates by propagating the outputs $f(x)$ backwards, subject to the conservation rule \cite{bach2015pixel}. Given neurons $j$ and $k$ in two consecutive layers, propagating relevance score $R_k$ to neurons $j$ in lower layer can be expressed as $R_j = \sum_k \frac{z_{jk}}{\sum_j z_{jk}} R_k $
 where weight $z_{jk} = w_{jk}x_k$ is the weighted activation, representing the contribution of relevance neuron $k$ makes to neuron $j$. 

 \paragraph{DeepLift:} It is an improved version of LRP by considering changes in the neuron activation from the reference point when propagating the relevance scores \cite{shrikumar2017learning}. Rescale rule is used to assign contribution scores to each neuron.

 \paragraph{Perturbation-based:}  LIME trains an interpretable local surrogate model, such as liner regression model, by sampling points around the input sample and use the regression coefficients as interpretation results \cite{ribeiro_why_2016}. SHAP calculates the attribution based on Shapley Values from cooperative game theory \cite{lundberg2017unified}. It involves taking the permutation of input features and adding them one by one to the baseline. The output difference after adding input feature corresponds to its attribution. 

\subsection{Classification and Interpretation Robustness}
\label{classification_vs_interpretation}

Suppose the gradient based interpretation can be written as $g(x) = \nabla \ell (x)$, where $\ell$ can be the cross-entropy loss (or our defined prediction loss $J$). We leverage Lipschitz continuous gradient to hint the relation between classification robustness and interpretation robustness as what follows.

A differentiable function $\ell (x)$ is called smooth within local region $B(x,r)$ iff it has a Lipschitz continuous gradient, i.e., if $\exists K > 0$ such that
\begin{equation}
     ||\nabla \ell(x')- \nabla \ell(x)|| \leq K ||x'-x||, \quad \forall x' \in B(x,r).
\end{equation}

\begin{prop}
\label{prop_lpg}
Lipschitz continuous gradient implies:
\begin{equation}
    ||\ell(x')-\ell(x)||  \leq ||\nabla \ell(x)|| r + \frac{K}{2} r^2
\end{equation}
\end{prop}
Prop.~\ref{prop_lpg} says, the change of classification is bounded by input gradient $||\nabla \ell(x)||$, as well as $\frac{K}{2}$. $K$ can be chosen as the Frobenius norm of input hessian $||H||_F(x)$ \cite{dombrowski2022towards}. Therefore, the regularisation of input gradient and input hessian can affect classification robustness and interpretation robustness.
\paragraph{Proof.} We first show that for $K >0$, $||\nabla \ell(x_1)- \nabla \ell(x_2)|| \leq K ||x_1-x_2||$ implies
$$
\ell(x_1) - \ell(x_2) \leq \nabla \ell(x_2)^T(x_1-x_2) + \frac{K}{2}||x_1-x_2||^2
$$
Recall from the integral calculus $\ell(a) - \ell(b) = \int_b^a \nabla\ell(\theta) \, d\theta$,
\begin{align*}
 &\ell(x_1)-\ell(x_2) =\\
 &\int_{0}^{1} \nabla \ell(x_2 + \tau (x_1-x_2))^T (x_1-x_2) \, d\tau =\\
 &\int_{0}^{1} (\nabla \ell(x_2 + \tau (x_1-x_2))^T -\nabla \ell(x_2)^T +\nabla \ell(x_2)^T ) \\
 &(x_1-x_2) \, d\tau 
\end{align*}
As $\nabla \ell(x_2)$ is independent of $\tau$, it can be taken out from the integral
\begin{align*}
    \ell(x_1) - \ell(x_2) = \nabla \ell(x_2)^T(x_1-x_2) + \\
    \int_{0}^{1} (\nabla \ell(x_2 + \tau (x_1-x_2))^T -\nabla \ell(x_2)^T) (x_1-x_2) \, d\tau 
\end{align*}
 Then we move $\nabla \ell(x_2)^T(x_1-x_2)$ to the left and get the absolute value
 \begin{align*}
     |\ell(x_1) - \ell(x_2) - \nabla \ell(x_2)^T(x_1-x_2)| = \\
     |\int_{0}^{1} (\nabla \ell(x_2 + \tau (x_1-x_2))^T -\nabla \ell(x_2)^T) (x_1-x_2) \, d\tau| \leq \\
     \int_{0}^{1} |(\nabla \ell(x_2 + \tau (x_1-x_2))^T -\nabla \ell(x_2)^T) (x_1-x_2)| \, d\tau \leq_{c.s.} \\
     \int_{0}^{1} ||(\nabla \ell(x_2 + \tau (x_1-x_2)) -\nabla \ell(x_2))|| ||(x_1-x_2)|| \, d\tau
 \end{align*}
c.s. means Cauchy – Schwarz inequality. By applying lipschitz continuous gradient, we can get
\begin{align*}
    ||(\nabla \ell(x_2 + \tau (x_1-x_2)) -\nabla \ell(x_2))|| \\
    \leq K ||\tau(x_1-x_2)|| \\
    \leq K \tau ||x_1-x_2||
\end{align*}
Note $\tau\geq0$, and the absolute sign of $\tau$ can be removed. Then, we can get
\begin{align*}
    |\ell(x_1) - \ell(x_2) - \nabla \ell(x_2)^T(x_1-x_2)| \leq \\
    \int_{0}^{1} K \tau ||x_1-x_2||^2 \, d\tau = \frac{K}{2}||x_1-x_2||^2
\end{align*}
Next, get the norm of two sides, and apply triangle inequality, we finally get
\begin{equation}
\begin{split}
    ||\ell(x')-\ell(x)|| & \leq || \nabla \ell(x)^T(x'-x) + \frac{K}{2}||x'-x||^2|| \\
    & \leq ||\nabla \ell(x)|| ||x'-x|| + \frac{K}{2}||x'-x||^2 \\
    & \leq ||\nabla \ell(x)|| r + \frac{K}{2} r^2 \\
\end{split}
\end{equation}
\textbf{QED}

\subsection{Genetic Algorithm based Optimisation}
\label{appendix_GA}
Genetic Algorithm (GA) is a classic evolutionary algorithm for solving the either constrained or unconstrained optimisation problems. It mimics the biological evolution by selecting the most fitted individuals in the population, which will be the parents for the next generation. It consists of 4 steps: initialisation, selection, crossover, and mutation, the last three of which are repeated until the convergence of fitness values.

\paragraph{Initialisation} The initialisation of population is crucial to the quick convergence. Diversity of initial population could promise approximate global optimal\cite{konak2006multi}. Normally, we use the Gaussian distribution with the mean at input seed $x$, or a uniform distribution to generate a set of diverse perturbed inputs within the norm ball $B(x,r)$.

\paragraph{Selection} A fitness function is defined to select fitted individuals as parents for the latter operations. We use the fitness proportionate selection \cite{lipowski2012roulette}. 
\begin{equation}
    p_i = \frac{\mathcal{F}_i}{\sum_{i=1}^{n} \mathcal{F}_i} 
\end{equation}

The fitness value is used to associate a probability of selection $p_i$ for each individuals to maintaining good diversity of population and avoid premature convergence. The fitness function is the objective function to be optimised. For example, previous paper applies GA to the perturbation optimisation to generate the high quality AEs \cite{DBLP:journals/compsec/ChenSSXZ19}. In this paper, the explanation discrepancy is optimised to find the worst case adversarial explanations.

\begin{figure}[!htb]
\centering
\includegraphics[width=\linewidth]{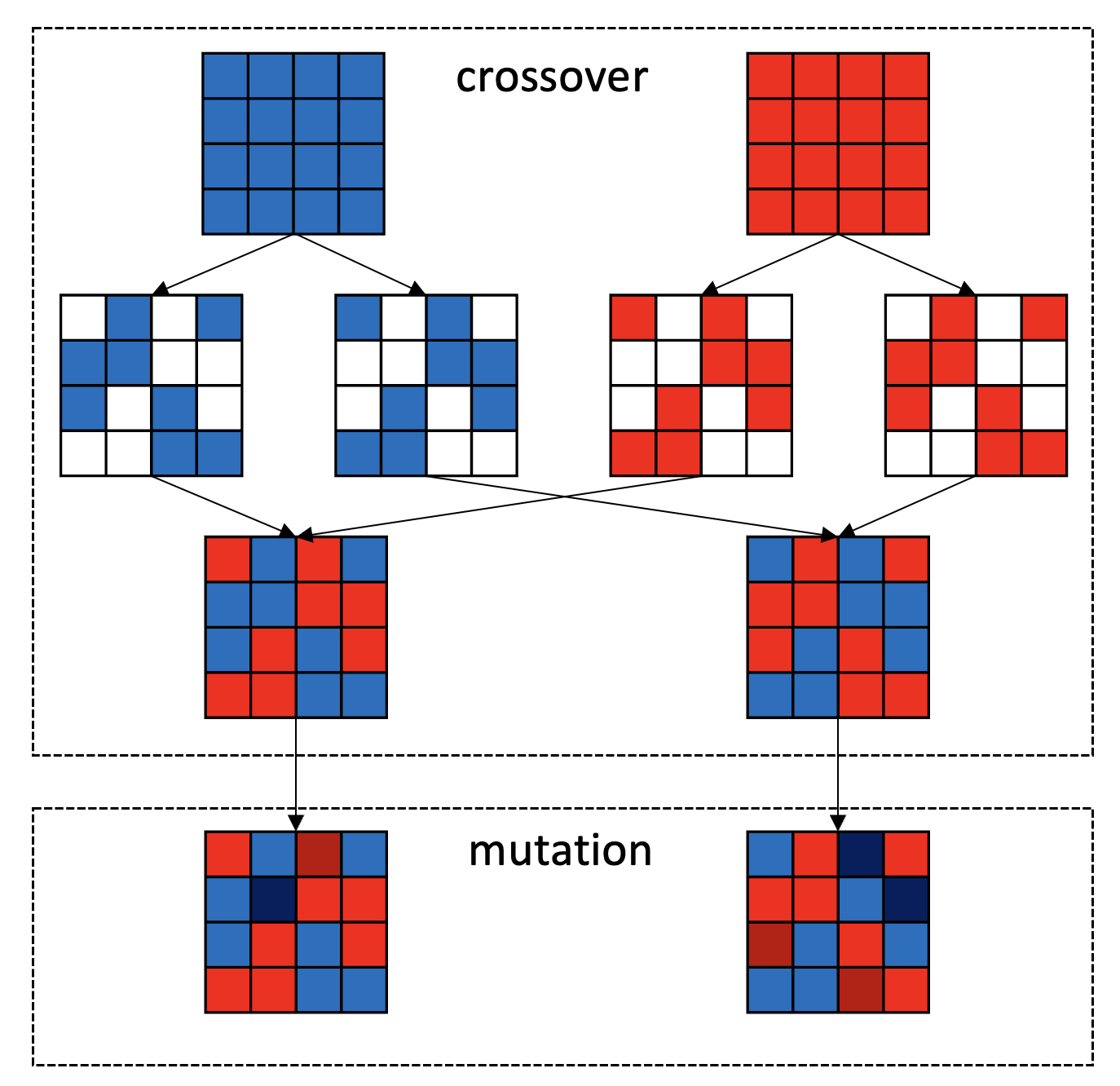}
\caption{Illustration of crossover and mutation in GA}
\label{fig:ga_crossover_mutation}
\end{figure}

\paragraph{Crossover} The crossover operator will combine a pair of parents from last step to generate a pair of children, which share many of the characteristics from the parents. The half elements of parents are randomly exchanged.

\paragraph{Mutation} Some elements of children are randomly altered to add variance in the evolution. It should be noticed that the mutated samples should still fall into the norm ball $B(x,r)$. Finally, the children and parents will be the individuals for the next generation.

\paragraph{Termination} The termination condition of GA is either maximum number of iterations is reached or the highest ranking of fitness reaches a plateau such that successive iterations no longer produce better results. In this paper, we fix the maximum iteration number for simplicity.

GA can be directly applied to the unconstrained optimisation when objective function equals to fitness function. The constraint optimisation is more challenging and different strategies are proposed to handle the non-linear constraint for GA \cite{michalewicz1996evolutionary}. One of the popular approaches is based on the superiority of feasible individuals to make distinction between feasible and infeasible solutions \cite{10.5555/645513.657601}. 

\subsection{Subset Simulation}
\label{appendix_ss}
Subset Simulation (SS) is widely used in reliability engineering to compute the small failure probability. The main idea of SS is introducing intermediate failure events so that the failure probability can be expressed as the product of larger conditional failure probabilities \cite{au2001estimation}.

Suppose the distribution of perturbed inputs with the norm ball is $q(x)$, and the failure event is denoted as $F$. let $F=F_m \subset F_{m-1} \subset \cdots \subset F_2 \subset F_1$ be a sequence of increasing events so that $F_m = \bigcap_{i=1}^{m}F_i$. By the definition of conditional probability, we get
\begin{equation}
\label{eq:subset_simulation_appendix}
\begin{split}
    P_F & = P(F_m) = P(\bigcap_{i=1}^{m}F_i) \\
        & = P(F_m | \bigcap_{i=1}^{m-1}F_i)P(\bigcap_{i=1}^{m-1}F_i) \\
        & = P(F_m | F_{m-1})P(\bigcap_{i=1}^{m-1}F_i) \\
        & = P(F_m | F_{m-1}) \cdots P(F_2 | F_1) P(F_1) \\
        & = P(F_1)\prod_{i=2}^{m}P(F_{i}|F_{i-1})
\end{split}
\end{equation}
$F_m$ is usually a rare event, which means a large amount of samples are required for the precise estimation by Simple Monte Carlo (SMC). SS decomposes the rare event with a series of intermediate events, which are more frequent. The conditional probabilities of intermediate events involved in Eq.~\eqref{eq:subset_simulation} can be chosen sufficiently large so that they can be efficiently estimated. For example, $P(F_1)=1$, $P(F_{i}|F_{i-1})=0.1$, $i=2,3,4,5,6$, then $P_F \approx 10^{-5}$ is too small for the efficient estimation by SMC.

The keypoint of SS is estimating $P(F_1)$ and conditional probabilities $P(F_{i}|F_{i-1})$. On the one hand, $F_1$ can be chosen as the common event such that by SMC of $N$ perturbed inputs within the norm ball $x'_k \sim q(x')$, all samples fall into $F_1$. On the other hand, computing the conditional probability \begin{equation}
\label{eq:pf1}
    P(F_{i+1}|F_{i}) = \frac{1}{N} \sum_{k=1}^N \mathbbm{1}_{F_{i+1}}(x'_k) \approx \rho
\end{equation} 
requires the simulation of $(1-\rho) N$ additional samples. For example, if we have $N$ samples belonging to $F_{i-1}$ with $i\geq2$, and $P(F_i|F_{i-1}) = \rho$, which indicate $\rho N$ samples belongs to $F_i$. To estimate next conditional probability $P(F_{i+1}|F_i)$, $(1-\rho) N$ additional samples lying in $F_i$ should be simulated to expand the population size to $N$. Given the conditional distribution $q(x'|F_i) = q(x')I_{F_1}(x')/P(F_i)$, on average $1/P(F_i)$ samples are simulated before one such sample occur. The Markov Chain Monte Carlo based on Metropolis-Hastings (MH) algorithm can be adopted to improve the efficiency.

At intermediate iteration $i$, we already obtain $\rho N$ samples lying in $F_i$, that is $x' \in F_i$. The target distribution is $q(\cdot|F_i)$. We can use MH algorithm to generate new samples $x''$ from the proposal distribution $g(x''|x')$. $g(x''|x')$ can be normal distribution or uniform distribution centred at $x'$.
The MH algorithm can be written as below:
\subsubsection{Initialisation} Pick up a sample $x'$ belonging to $F_i$. Set step $t = 0$ and let $x_t = x'$.
\subsubsection{Iteration} At step $t$, generate a random candidate sample $x''$ according to $g(x''|x_t)$. Calculate the acceptance probability
\begin{equation}
    A(x'',x_t) = \min\{1,\frac{q(x_t|F_i)}{q(x''|F_i)}\frac{g(x_t|x'')}{g(x''|x_t)}\}
\end{equation}
and accept the new sample $x''$ with probability $A(x'',x_t)$. Further check if $x''\in F_i$, otherwise reject $x''$. In practice, we generate a uniform random number $u \in [0,1]$, set $x_{t+1}$ as
\begin{equation}
    x_{t+1} = 
    \begin{cases}
      x'' & \text{if $u \leq A(x'',x_t)$ and $x'' \in F_i$}\\
      x_t & \text{Otherwise}\\
    \end{cases} 
\end{equation}
and increment $t = t+1$. 

We can run a large amount of Markov chains simultaneously to enlarge the set of i.i.d. samples falling into $F_i$. However, as discussed in \cite{katafygiotis2008geometric,schueller2004critical}, MH becomes inefficient for high dimensional problems. The acceptance probability $A(x'',x')$ will rapidly decrease with increasing dimensions. It results in many repeated samples and high correlated Markov chains. It is recommended to adapt the proposal distribution $g(x''|x')$ after $M$ steps of MH \cite{papaioannou2015mcmc}. The mean acceptance probability should be kept around 0.234 \cite{gelman1997weak}.

The whole process of SS can be summarized as follows. First, we simulate $N$ perturbed samples within the norm ball $B(x,r)$ (all belong to $F_1$) and use SMC to estimate $P(F_2|F_1)$. From these $N$ samples, we already obtain $\rho N$ samples distributed from $q(\cdot|F_2)$. Start from each of these $\rho N$ samples falling in $F_2$, we can create a Markov chain and run MH $M$ steps to generate new samples distributed from $q(\cdot|F_2)$. In initial SS method \cite{au2001estimation}, $\rho N$ distinct Markov chains (with different start points) are created. $1/\rho$ new samples are drawn from each chain, and the covariance between new samples in same Markov chain should be considered for evaluating the coefficient of variation (c.o.v) of the final estimation on ${P}_F$. \cite{cerou2012sequential} modify the algorithm by firstly enlarge set to $N$ samples with replacement from $\rho N$. Then $N$ Markov Chains are constructed and only one sample is drawn from each chain. 

These new generated samples can be utilised to estimate $P(F_3|F_2)$. Repeating this process
until the rare failure of interest. We get the final estimation of failure event probability by ``assembling'' the conditional probabilities with Eq.~\eqref{eq:subset_simulation}.

\subsubsection{Statistical Property of SS Estimator} We present the analysis on statistical property of $P_{F_i}$ (shortened notation for $P(F_1)$ and $P(F_{i}|F_{i-1})$) and $P_F$. They are based on the assumption that Markov chain generated by MH algorithm is theoretically ergodic. That is, the stationary distribution is unique and tend to the corresponding conditional probability distribution. Since we simulate samples from Markov chain to estimate $P_{F_i}$ (ref. to Eq.~\eqref{eq:pf1}), The coefficient of variation of $P_{F_i}$ (c.o.v) is 
\begin{equation}
\delta_i = \sqrt{\frac{1-P_{F_i}}{P_{F_i} N}}(1+\lambda_i)     
\end{equation}

$\lambda_i >0$ represents the dependency of samples drawn from Markov Chain. This is compared to case when we use SMC to simulate independent samples from the known distribution ($\lambda_i=0$). As $N\rightarrow\infty$, the Central Limit Theorem (CLT) tells $\widebar{P}_{F_1}\rightarrow{P}(F_1)$, and $\widebar{P}_{F_i}\rightarrow{P}(F_{i}|F_{i-1})$. We can get almost surely $\widebar{P}_F\rightarrow P(F_1)\prod_{i=2}^{m}P(F_{i}|F_{i-1})=P_F$. It should be noted that $\widebar{P}_{F}$ is biased for $N$, but asymptotically unbiased due to the fact that samples in $F_i$ for computing $\widebar{P}_{F_i}$ are utilised to start Markov chain for computing $\widebar{P}_{F_{i+1}}$. This bias will asymptotically vanish when $N$ goes to infinity.
\begin{prop}
\label{prop_pf_mean}
$\widebar{P}_{F}$ is biased for $N$, the fractional bias is bounded by:
\begin{equation}
\label{ss_bias}
    |E\left[\frac{\widebar{P}_{F}-P_{F}}{P_{F}}\right]| \leq \sum_{i>j} \delta_i \delta_j + o(1/N) = O(1/N)
\end{equation}
\end{prop}

\paragraph{Proof.} We define $Z_i = (\widebar{P}_{F_i}-P_{F_i})/\sigma_i$, and get $\widebar{P}_{F_i}=P_{F_i}+\sigma_iZ_i$. By CLT, it's clear that $E[Z_i]=0$ and $E[Z^2_i]=1$.
\begin{align*}
\frac{\widebar{P}_{F}-P_{F}}{P_{F}} &= \prod_{i=1}^{m}\widebar{P}_{F_i}/P_{F_i} - 1 \\
&= \prod_{i=1}^{m} (1+\delta_iZ_i) - 1 \\
&= \prod_{i=1}^{m}\delta_i Z_i + \sum_{i=1}^{m} \delta_i Z_i + \sum_{i>j} \delta_i \delta_j Z_i Z_j + \\
&\sum_{i>j>k} \delta_i \delta_j \delta_k Z_i Z_j Z_k + ...
\end{align*}
Take expectation and use $E[Z_i]=0$, we can further get 
\begin{align*}
E \left[\frac{\widebar{P}_{F}-P_{F}}{P_{F}}\right] &= \left(\prod_{i=1}^{m} \delta_i\right) E\left[\prod_{i=1}^m Z_i\right] + \sum_{i>j} \delta_i \delta_j E[Z_i Z_j] \\
&+ \sum_{i>j>k} \delta_i \delta_j \delta_k E[Z_i Z_j Z_k] + ...
\end{align*}
Since $\{Z_i\}$ are correlated, $E[Z_iZ_j]$, $E[Z_i Z_j Z_k]$,.... are not zero, and $\widebar{P}_{F_i}$ is biased for every $N$. $\delta_i$ is $O(1/\sqrt{N})$ according to the definition, which makes $\sum_{i>j} \delta_i \delta_j E[Z_i Z_j]$ have $O(1/N)$ and remaining items with higher product of $\delta_i$ have $o(1/N)$. Take absolute value of both sides and use Cauchy-Schwartz inequality to obtain $|E[Z_iZ_j]|\leq\sqrt{E[Z^2_i]E[Z^2_j]} = 1$. Finally, we can get the proof.

\begin{prop}
\label{prop_pf_cov}
$\widebar{P}_{F}$ is a consistent estimator and its c.o.v. $\delta$ is bounded by:
\begin{equation}
\label{ss_cov}
    \delta^2 = E\left[\frac{\widebar{P}_{F}-P_{F}}{P_{F}}\right]^2 \leq \sum_{i,j=1} \delta_i \delta_j + o(1/N) = O(1/N)
\end{equation}
\end{prop}

\paragraph{Proof.}
\begin{align*}
& E\left[\frac{\widebar{P}_{F}-P_{F}}{P_{F}}\right]^2 \\
&=E\left[\prod_{i=1}^{m}\delta_i Z_i + \sum_{i=1}^{m} \delta_i Z_i + \sum_{i>j} \delta_i \delta_j Z_i Z_j +...\right]^2 \\
&= \sum_{i,j=1}^m \delta_i \delta_j E[Z_i Z_j] + o(1/N) \\
&\leq \sum_{i,j=1}^m \delta_i \delta_j + o(1/N) = O(1/N)
\end{align*}
As $\delta_i = O(1/\sqrt{N})$ and $E[Z_iZ_j]\leq 1$. Note that the bias is accounted for when c.o.v. $\delta$ is defined as the deviation about $P_{F}$, instead of $E[\widebar{P}_{F}]$. The upper bound corresponds to the case that conditional probability $\{P_{F_i}\}$ are all correlated. Although $\{P_{F_i}\}$ are generally correlated, $\delta$ can be well approximated by $\sum_{i=1}^m \delta^2_i$. For simplicity, we can also make the assumption that enough steps of MH algorithm are taken to eliminate the dependency of simulated samples from MCMC ($\lambda_i=0$) \cite{cerou2012sequential}. Then we use sample mean $\widebar{P}_{F_i}$ to approximate $P_{F_i}$, and finally get
\begin{equation}
\widebar{\delta}^2 \approx \sum_{i=1}^m \delta^2_i = \sum_{i=1}^m \frac{1-\widebar{P}_{F_i}}{\widebar{P}_{F_i} N}(1+\lambda_i) \approx \sum_{i=1}^m \frac{1-\widebar{P}_{F_i}}{\widebar{P}_{F_i} N}  
\end{equation}
To get an idea of how many samples are required by SS to achieve the estimation accuracy $P_F$, we assume the c.o.v $\delta$, $\lambda_i=\lambda$ and $P(F_{i}|F_{i-1})=\rho$ are fixed, then $m=logP_F/log\rho+1$, and $\delta^2=(m-1)\frac{1-\rho}{\rho N}(1+\lambda)$, We can get the number of simulated samples in SS is
$$
N_{SS} \approx mN = (\frac{|logP_F|^2}{|log\rho|^2}+\frac{|logP_F|}{|log\rho|})\frac{(1-\rho)(1+\lambda)}{N\delta^2}
$$
Thus, for a fixed $\delta$ and $\rho$, $N_{SS}\propto (|logP_F|^2+|log\rho||logP_F|)$. Compared to the SMC, 
 the required samples are $N_{SMC}\propto1/P_F$. This indicates that SS is substantially efficient to estimate small failure probability.


\subsection{Complexity Analysis of Genetic Algorithm and Subset Simulation Applied on XAI Methods}

Although the proposed evaluation methods can be applied to all kinds of feature attribution based XAI techniques, the time complexity will be extremely high for perturbation based XAI methods, such as LIME and SHAP, which take random perturbation of input features to yield explanations.

The complexity of GA is $O(t\cdot N \cdot(c(fitness) + c(crossover)+ c(mutation) ))$, where $t$ and $N$ are evolution iterations and population size, respectively. When we choose different XAI methods, the evaluation time of fitness values $c(fitness)$ will change correspondingly.

The complexity of SS is related to the number of sub-events $m$, the number of MH steps $M$ and number of simulated samples $N$. For estimating conditional probability of each sub-event, $M$ MH steps are taken, and running each MH step requires the calculation of property function of $N$ samples. Thus, the complexity of SS is approximately $O(m\cdot M \cdot N\cdot c(property))$.  When we choose different XAI methods, the evaluation time of property function $c(property)$ will change correspondingly.

\begin{table}[ht]
\centering
\caption{Time counts of $N\cdot c(cal\_attr\_dis)$ in seconds across different dataset ($N=1000$). Results are averaged over 10 runs.}
\resizebox{\linewidth}{!}{
\begin{tabular}{ccccccc}
\hline
Dataset & \begin{tabular}[c]{@{}c@{}}Gradient \\ x Input\end{tabular} & \begin{tabular}[c]{@{}c@{}}Integrated\\ Gradients\end{tabular} & GradCAM & DeepLift & LIME & SHAP \\ \hline
MNIST & 0.0202 & 0.0512 & 0.0342 & 0.0382 & 99.21 & 25.80 \\
CIFAR-10 & 0.0909 & 0.3329 & 0.1222 & 0.1307 & 293.72 & 255.95 \\
CelebA & 0.0620 & 0.2759 & 0.0887 & 0.1029 & 739.59 & 692.75 \\ \hline
\end{tabular}
}
\label{tab:complexity_analysis}
\end{table}

From the definition of fitness function in GA and property function in SS. both $c(fitness)$ and $c(property)$ can be approximated by the computation of interpretation discrepancy $c(cal\_attr\_dis)$. In practice, we can compute interpretation discrepancy in a batch, e.g. $N$ samples can run simultaneously to generate the explanations. Therefore, we count the running time of $N\cdot c(cal\_attr\_dis)$ across different datasets and different XAI methods in Nvidia A100. Results are presented in Table~\ref{tab:complexity_analysis}. LIME and SHAP take much more time than gradient-based XAI methods for the batch computation of interpretation discrepancy. This will be amplified by iteration number $t$ in GA or number of sub-events times number of MH steps $m \cdot M$ in SS for one time evaluation of interpretation robustness.

\subsection{Details of DL models}
\label{exp_model_details}
The information of DL models under evaluation are presented in Table~\ref{table_model_details}.
All experiments were run on a machine of Ubuntu 18.04.5 LTS x86\_64 with Nvidia A100 GPU and 40G RAM. The source code, DL models, datasets and all experiment results are available in Supplementary Material, and will be publicly accessible at GitHub after the double-blind review process.

\begin{table*}[ht]
\centering
\caption{Details of the datasets and DL models under evaluation.}
\resizebox{0.8\linewidth}{!}{
\begin{tabular}{cccccccccccc}
\hline
\multirow{2}{*}{Dataset} & \multirow{2}{*}{Image Size} & \multirow{2}{*}{$r$} & \multirow{2}{*}{DL Model} & \multicolumn{2}{c}{Org.} & \multicolumn{2}{c}{Grad. Reg.} &
\multicolumn{2}{c}{Hess. Reg.} & \multicolumn{2}{c}{Adv. Train.} \\
 & & & & Train & Test  & Train & Test & Train & Test & Train & Test \\ \hline
MNIST & $1\times32\times32$ & 0.1 & LeNet5 & $1.000$ & $0.991$ & $0.993$ & $0.989$ & $0.993$ & $0.989$ & $0.994$ & $0.989$ \\
CIFAR-10 & $3\times32\times32$ & 0.03 & ResNet20 & $0.927$ & $0.878$ & $0.910$ & $0.876$ & $0.786$ & $0.779$ & $0.715$ & $0.703$ \\
CelebA & $3\times64\times64$ & 0.05 & MobileNetV1 & $0.934$ & $0.917$ & $0.918$ & $0.912$ & $0.908$ & $0.904$ & $0.769$ & $0.789$ \\ \hline
\end{tabular}}
\label{table_model_details}
\end{table*}

\subsection{Experiment on Interpretation Discrepancy Measures}
\label{appendix_interpretation_discrepancy}

\begin{figure*}[!htbp]
\centering
\includegraphics[width=0.9\linewidth]{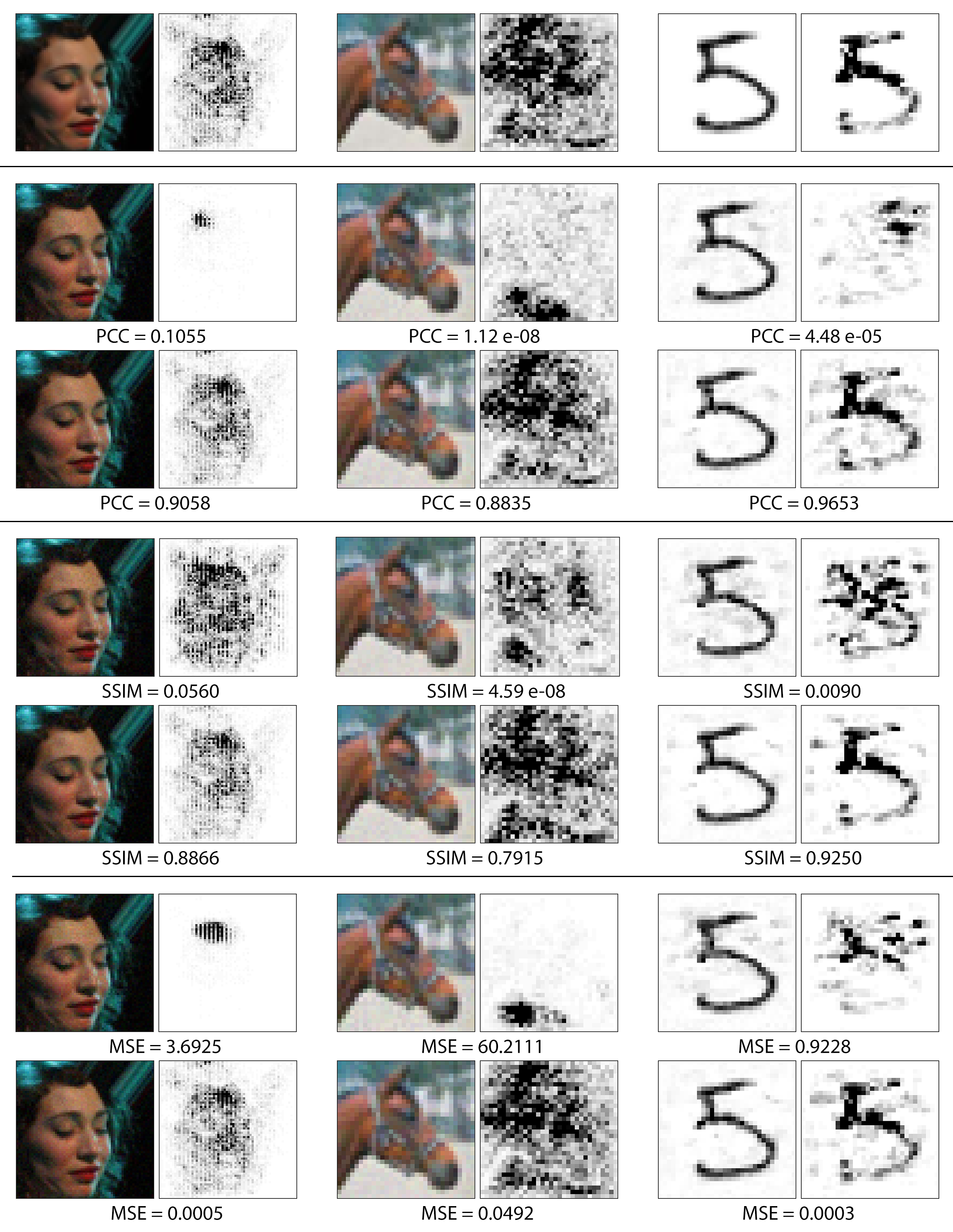}
\caption{Comparison between PCC, SSIM and MSE as metrics of interpretation discrepancy between original interpretation and adversarial interpretation, generated by GA and SS. Smaller PCC, smaller SSIM, and larger MSE indicate greater difference. In this set of experiments, PCC is relatively the best to quantify the visual difference between attribution maps.}
\label{fig:pcc_ssim_mse}
\end{figure*}

We study the quality of three widely used metrics, i.e. Mean Square Error (MSE), Pearson Correlation Coefficient (PCC), and Structural Similarity Index Measure (SSIM) \cite{dombrowski2019explanations} to quantify the visual discrepancy between two attribution maps. The proposed evaluation methods can produce the adversarial interpretation with the guidance of different metrics. As shown in Fig.~\ref{fig:pcc_ssim_mse}, the first row displays three seed inputs and corresponding attribution maps. The following groups separated by lines show the adversarial interpretation of perturbed input measured by different metrics. The value of PCC appears to be relatively more accurate in terms of reflecting the visual difference between original interpretation of seeds input and adversarial interpretations. Smaller PCC represents larger visual difference between two attribution maps. In addition, the value range of PCC is 0$\sim$1, with 0$\sim$0.3 indicating weak association, 0.5$\sim$1.0 indicating strong association. Therefore, it provides a uniform measurement across different seeds input and different dataset. In contrast, MSE can also precisely measure the visual difference but vary greatly with respect to seed inputs and image size. SSIM exhibits the worst performance in measuring difference between attribution maps.

\subsection{Experiment on Parameter Sensitivity}
\label{appendix_parameter}
\begin{figure*}[!ht]
\centering
\includegraphics[width=\linewidth]{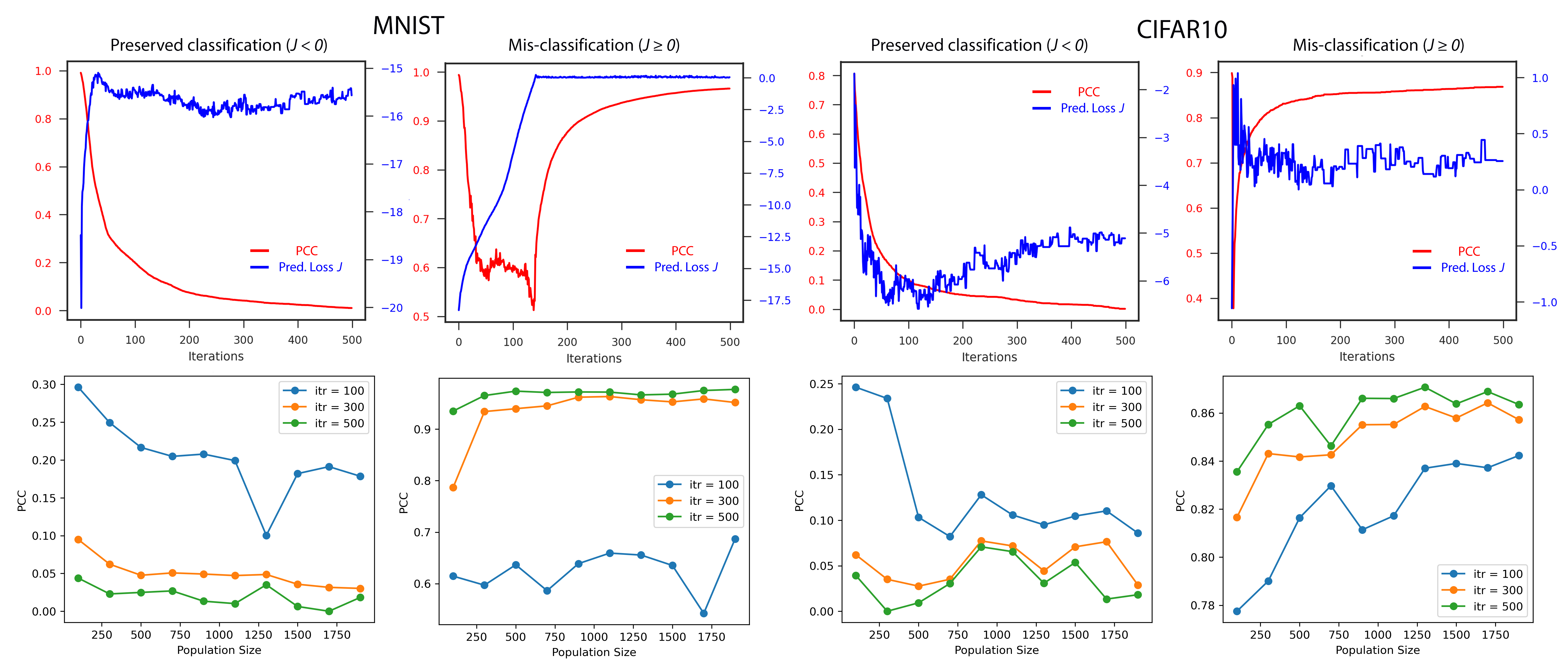}
\caption{GA is applied to test seeds (norm balls) from MNIST and CIFAR10 dataset to find worst case interpretation discrepancy, measure by PCC. First row: fixed population size 1000, and varied iterations; Second row: fixed iterations, and varied population size. ``Gradient$\times$Input'' interpretation method is considered.}
\label{fig:ga_accuracy_additional}
\end{figure*}

\begin{figure*}[!htbp]
\centering
\includegraphics[width=\linewidth]{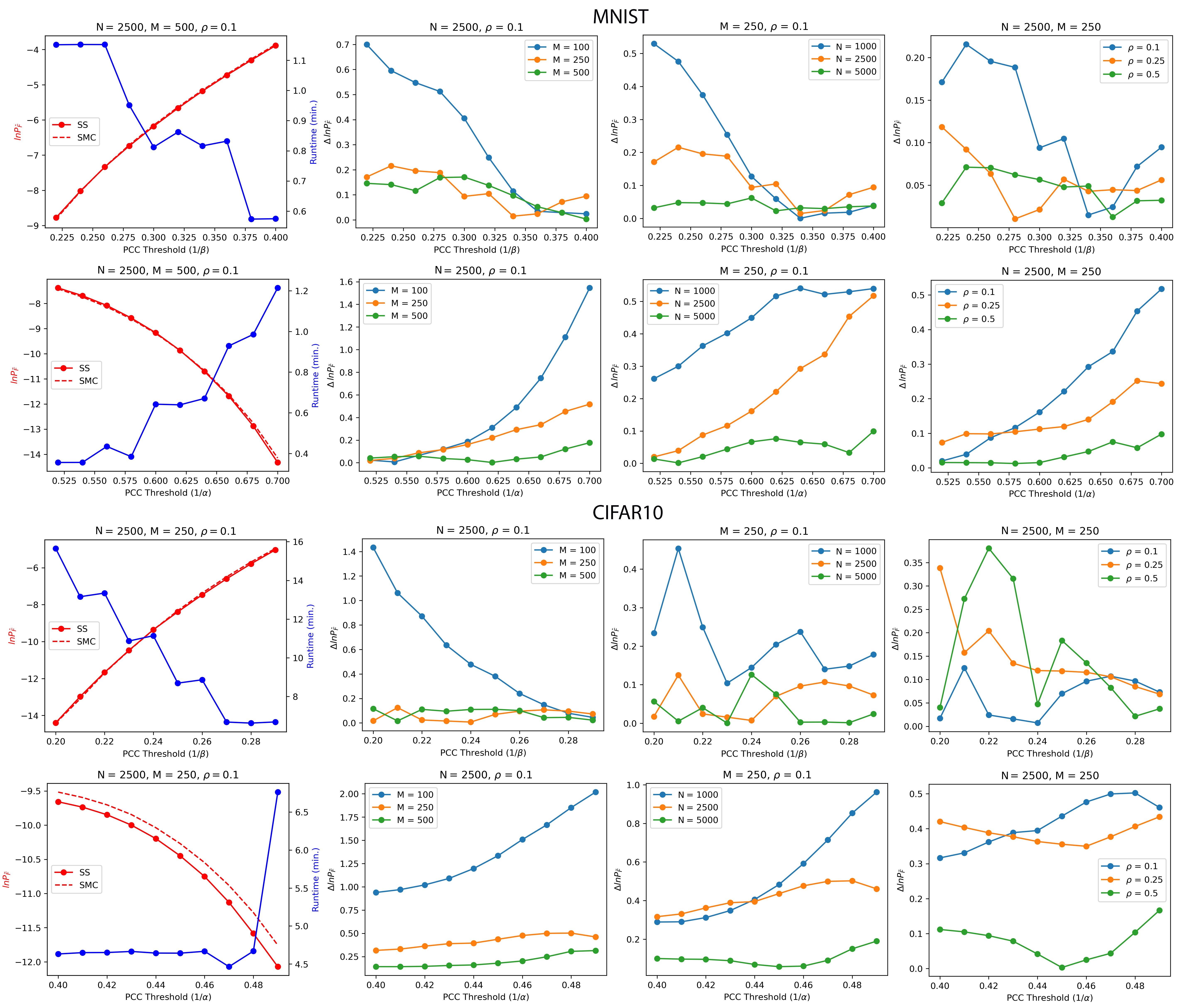}
\caption{SS for estimating the probability of misinterpretation ($\ln P_F$) within a norm ball from MNIST, CIFAR10 dataset compared with SMC using $10^8$ samples (~22 minutes for each estimate for MNIST; ~154 minutes for each estimate for CIFAR10). Results are averaged on 10 runs.  ``Gradient$\times$Input'' interpretation method is considered.}
\label{fig:ss_para}
\end{figure*}

Additional experiments on hyper-parameter settings in GA and SS are presented in Fig.~\ref{fig:ga_accuracy_additional} and Fig.~\ref{fig:ss_para}. The objective function interpretation discrepancy $\mathfrak{D}$, measured by PCC, is optimised to converge with the increasing number of iterations while the prediction loss $J$ as the constraint is gradually satisfied. The number of iterations
in GA is more important than population size.

For hyper-parameters in SS, apart from the sensitivity of MH steps, we also discuss the impact of population size $n$ and quantile $\rho$ for conditional probability. As expected, increasing population size will improve the estimation precision, using SMC results with $10^8$ samples as the ground truth. However, there is no exact answer for which $\rho$ is better. In most cases, we find that $\rho = 0.5$ can reduce the estimation error, but will take more time for one estimation. Larger $\rho$ represents more sub events are decomposed and additional estimation of conditional probability will obviously cost more time. Fortunately, we find SS estimation accuracy is more sensitive to the number of MH steps $M$ and population size $n$, compared with $\rho$. Therefore, setting $\rho =0.1$ but increasing MH steps and population size will get 
sufficiently accurate results. Finally, the rarity of failure events can determine the setting of these hyper-parameters. The estimating accuracy of more rare events, e.g. $\text{PCC} < 0.2$, is more sensitive to the theses parameters.

\subsection{Experiments on Evaluating XAI methods}
\label{appendix_eval_xai}
\subsubsection{Evaluation for Gradient-based XAI Methods}
\begin{figure*}[!htbp]
\centering
\includegraphics[width=0.8\linewidth]{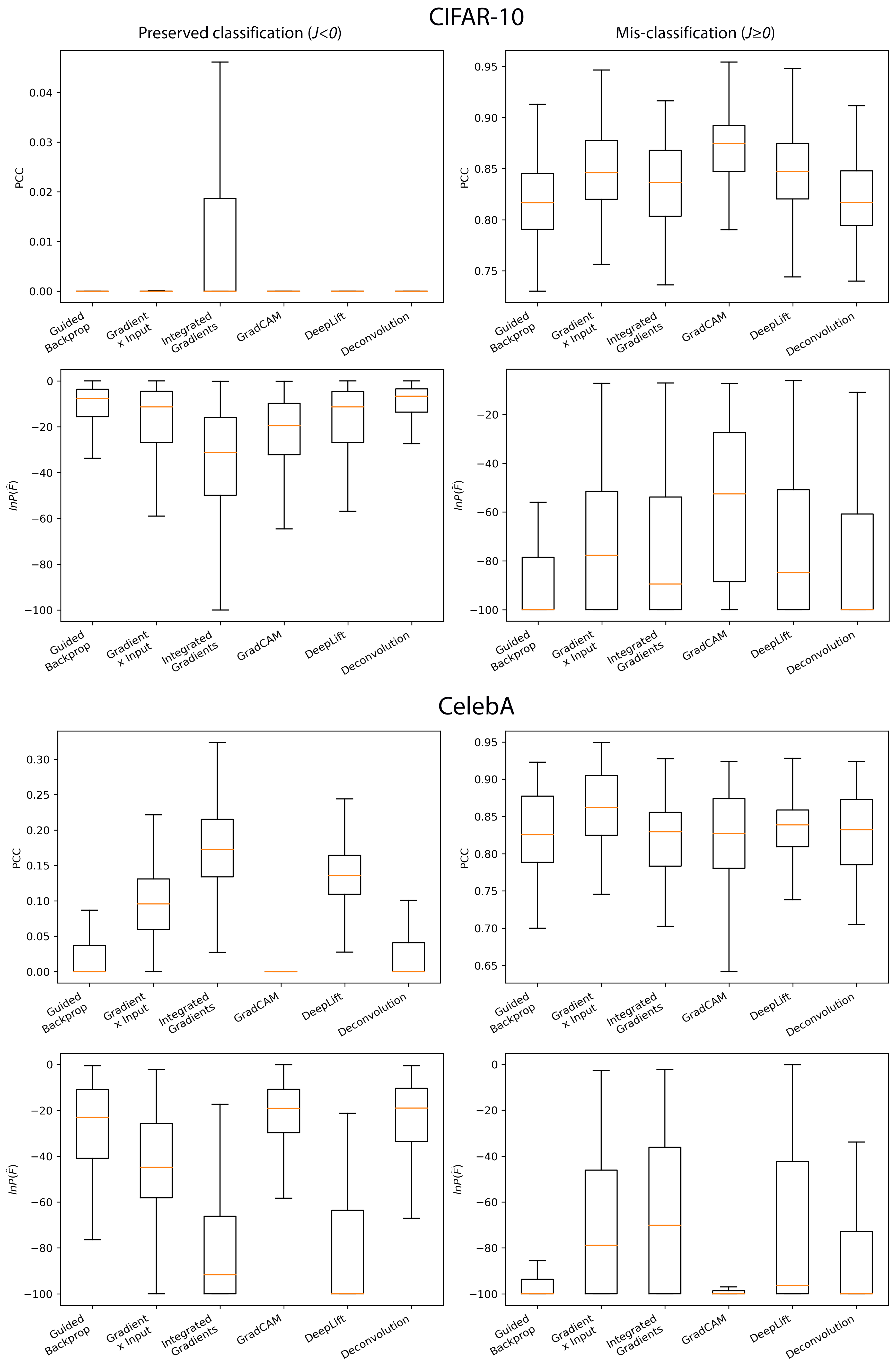}
\caption{Robustness evaluation of different interpretation methods based on 100 randomly selected samples from CIFAR10 and CelebA test set. From top to bottom, first row (worst case evaluation) and second row (probabilistic evaluation). From left to right, first column (misinterpretation $\widehat{F}$) and second column (misinterpretation $\widetilde{F}$)}
\label{fig:intepretation_eval_additional}
\end{figure*}

We evaluate the robustness of more XAI methods on CIFAR10 and CelebA dataset, including ``Deconvolution'', ``Guided Backpropagation'', ``Gradient$\times$Input'', ``Integrated Gradients'', ``GradCAM'', and ``DeepLift''. Results are presented in Fig.~\ref{fig:intepretation_eval_additional}. In terms of misinterpretation with preserved classification, Integrated Gradients is the most robust XAI method due to the integral of gradient of model's output with respect to the input. The integral averages the gradient-based attribution maps over several perturbed images instead of single point explanation. DeepLift has the similar smoothing mechanism by comparing the neuron activation with a reference point. Therefore, single point explanation like Deconvolution and GradCAM are vulnerable to this type of misinterpretation when DL model's loss surface is highly curved, leading to the great change of gradients. Gradient$\times$Input is slightly better by leveraging the input sign and strength.

These XAI methods in general show similar robustness against misinterpretation conditioned on misclassification, although we find the single point explanation is a litter better than explanation averaged over several points under this circumstance. We guess the rarity of misclassification and misinterpretation make it difficult to find the perturbed input which have different attribution map with input seeds. Therefore, the averaged interpretation of perturbed input tend to be consistent with original interpretation. 

\subsubsection{Evaluation for Perturbation-based XAI Methods}
We also consider the robustness of interpretation for LIME and SHAP, the most popular perturbation-based XAI methods. In contrast to the gradient-based XAI methods, the robustness problem of which is thoroughly studied, perturbation-based XAI methods are difficult to be attacked by adversarial noise due to the model-agnostic settings. As far as we have known, the only adversarial attack on LIME/SHAP \cite{slack2020fooling} requires to scaffold the biased DL model. That's conceptually different from the interpretation robustness mentioned in this paper, for which the internal structure of DL model should not be maliciously modified. Thanks to the black-box nature of our evaluation approaches, we can assess the robustness of LIME/SHAP. As is known, image feature segmentation is an important procedure in LIME/SHAP. LIME/SHAP will produce inconsistent interpretation at each run when the number of samples is smaller than the number of image segments \cite{zhao_baylime_2021}. Therefore, we record the evaluation results when using different number of samples. For simplicity, we use quickshift to segment the images into around 40 pieces of super-pixels, which is the default settings of LIME/SHAP tools.

\begin{table}[!htbp]
\centering
\caption{Robustness evaluation of perturbation-based XAI methods.}
\resizebox{\linewidth}{!}{
\begin{tabular}{c|c|cccc}
\hline
\multirow{2}{*}{Dataset} & \multirow{2}{*}{\begin{tabular}[c]{@{}c@{}}XAI Method\\ + Num\_Samples\end{tabular}} & \multicolumn{2}{c}{Worst Case Evaluation} & \multicolumn{2}{c}{Probabilistic Evaluation} \\ \cline{3-6} 
 &  & \begin{tabular}[c]{@{}c@{}}$sol_{\widehat{F}}$\\ (PCC)\end{tabular}  & \begin{tabular}[c]{@{}c@{}}$sol_{\widetilde{F}}$\\ (PCC)\end{tabular} & $\ln P_{\widehat{F}}$ & $\ln P_{\widetilde{F}}$  \\ \hline
\multirow{6}{*}{MNIST} & LIME+50 & 0.0002 & 0.9886 & -0.46 & -12.96 \\
 & LIME+200 & 6.88e-05 & 0.9350 & -0.37 & -14.59 \\ 
 & LIME+500 & 8.59e-06 & 0.8360 & -0.31 & -16.98 \\ \cline{2-6}
 & SHAP+50 & 4.11e-05 & 0.9648 & -0.36 & -14.78\\ 
 & SHAP+200 & 0.0011 & 0.9708 & -0.39 & -14.44 \\
 & SHAP+500 & 0.0005 & 0.9851 & -0.34 & -14.41 \\\hline
\multirow{6}{*}{CIFAR-10} & LIME+50 & 0.0002 & 0.9940 & -3.58 & -28.96 \\
 & LIME+200 & 0.0001 & 0.9986 & -3.78 & -30.28 \\ 
 & LIME+500 & 0.0001 & 0.9965 & -4.29 & -40.06 \\ \cline{2-6}
 & SHAP+50 & 0.0014 & 0.9973 & -3.75 & -48.56\\ 
 & SHAP+200 & 0.0016 & 0.9950 & -3.94 & -47.87 \\
 & SHAP+500 & 0.0001 & 0.9982 & -3.84 & -46.24 \\\hline
 \multirow{6}{*}{CelebA} & LIME+50 & 0.0004 & 0.9571 & -1.17 & -39.63 \\
 & LIME+200 & 1.23e-05 & 0.9824 & -4.06 & -41.41 \\ 
 & LIME+500 & 0.0001 & 0.9739 & -5.53 & -48.55 \\ \cline{2-6}
 & SHAP+50 & 0.0008 & 0.9568 & -4.24 & -49.21\\ 
 & SHAP+200 & 0.0006 & 0.9520 & -4.97 & -50.69 \\
 & SHAP+500 & 0.0002 & 0.9543 & -4.41 & -58.18 \\\hline
\end{tabular}
}
\label{xai_pert}
\end{table}

The initial results in Table~\ref{xai_pert} give us the hints that perturbation-based XAI methods also suffer from the lack of interpretation robustness, especially when classification is preserved but interpretation is different. In addition, increasing the number of perturbed samples is not significant to improving interpretation robustness. In other words, even if we use enough number of perturbed samples for LIME/SHAP to produce precise interpretation results, they are still easily fooled by adversarial noise. In the second experiment, we further explore the influence of image segmentation on interpretation robustness. By making the assumption that image segmentation is fixed or not fixed after adding adversarial noise, we can check whether adversarial noise change the image segmentation and indirectly affect the interpretation robustness of perturbation-based XAI methods. Result in Table~\ref{xai_pert_feature_segmentation} shows that current image segmentation used by LIME/SHAP is sensitive to the pixel-level adversarial noise and will produce different feature masks, which may affect the interpretation robustness. Nevertheless, fixing image segmentation is not effective to defend second type of misinterpretation-wrong classification with persevered interpretation.

\begin{table}[!htbp]
\centering
\caption{Sensitivity of Image Segmentation to adversarial noise when evaluating interpretation robustness for LIME+200.}
\resizebox{\linewidth}{!}{
\begin{tabular}{c|c|cccc}
\hline
\multirow{2}{*}{Dataset} & \multirow{2}{*}{\begin{tabular}[c]{@{}c@{}}Image\\ Segmentation\end{tabular}} & \multicolumn{2}{c}{Worst Case Evaluation} & \multicolumn{2}{c}{Probabilistic Evaluation} \\ \cline{3-6} 
 &  & \begin{tabular}[c]{@{}c@{}}$sol_{\widehat{F}}$\\ (PCC)\end{tabular}  & \begin{tabular}[c]{@{}c@{}}$sol_{\widetilde{F}}$\\ (PCC)\end{tabular} & $\ln P_{\widehat{F}}$ & $\ln P_{\widetilde{F}}$  \\ \hline
\multirow{2}{*}{MNIST} & Not Fixed & 6.88e-05 & 0.9350 & -0.37 & -14.59 \\
 & Fixed & 0.3632 & 0.8892 & -34.22 & -17.38\\ \hline
\multirow{2}{*}{CIFAR-10} & Not Fixed & 0.0001 & 0.9986 & -3.78 & -30.28 \\
 & Fixed & 0.0004 & 1.0000 & -100 & -41.33 \\\hline
 \multirow{2}{*}{CelebA} & Not Fixed & 1.23e-05 & 0.9824 & -4.06 & -41.41  \\
 & Fixed & 0.3547 & 0.8289 & -100 & -38.72\\ \hline
\end{tabular}
}
\label{xai_pert_feature_segmentation}
\end{table}

The above observations align with the insight that interpretation robustness is attributed to the geometrical properties of DL model (i.e. large curvature of loss function), but not the XAI methods. Therefore, the most effective way to address the problem is to train a DL model, which is more robust to be interpreted.

\begin{table}[!htbp]
\centering
\caption{Robustness evaluation of XAI methods on different neural network architecture for CIFAR-10 dataset.}
\resizebox{\linewidth}{!}{
\begin{tabular}{cccccc}
\hline
\begin{tabular}[c]{@{}c@{}}Model \\ Architecture\end{tabular} & \begin{tabular}[c]{@{}c@{}}Eval\\ Metrics\end{tabular} & \begin{tabular}[c]{@{}c@{}}Gradient \\ x Input\end{tabular} & \begin{tabular}[c]{@{}c@{}}Integrated\\ Gradients\end{tabular} & GradCAM & DeepLift \\ \hline
\multirow{4}{*}{ResNet20} & $sol_{\widehat{F}}$ & 0.0166 & \textbf{0.0375} & 0.0044 & 0.0212 \\
 & $sol_{\widetilde{F}}$ & 0.8562 & 0.8308 & \textbf{0.8079} & 0.8551 \\
 & $\ln P_{\widehat{F}}$ & -20.32 & \textbf{-45.05} & -35.93 & -21.22 \\
 & $\ln P_{\widetilde{F}}$ & -80.73 & \textbf{-87.64} & -68.27 & -81.81 \\ \hline
\multirow{4}{*}{MobileNetV2} & $sol_{\widehat{F}}$ & 0.0552 & \textbf{0.1167} & 0.0523 & 0.0712 \\
 & $sol_{\widetilde{F}}$ & 0.7689 & 0.7885 & \textbf{0.7085} & 0.7707 \\
 & $\ln P_{\widehat{F}}$ & -12.75 & \textbf{-34.99} & -16.01 & -8.70 \\
 & $\ln P_{\widetilde{F}}$ & -70.32 & -62.19 & \textbf{-82.17} & -68.38 \\ \hline
\multirow{4}{*}{VGG16} & $sol_{\widehat{F}}$ & 0.0767 & \textbf{0.1227} & 0.1133 & 0.0206 \\
 & $sol_{\widetilde{F}}$ & \textbf{0.7813} & 0.8240 & 0.8637 & 0.8358 \\
 & $\ln P_{\widehat{F}}$ & -14.42 & \textbf{-53.48} & -47.52 & -44.25 \\
 & $\ln P_{\widetilde{F}}$ & -59.74 & -54.155 & -49.90 & \textbf{-66.02} \\ \hline
\multirow{4}{*}{DLA} & $sol_{\widehat{F}}$ & 0.0737 & \textbf{0.0953} & 0.0078 & 0.0930 \\
 & $sol_{\widetilde{F}}$ & 0.7919 & 0.8111 & \textbf{0.2113} & 0.7983 \\
 & $\ln P_{\widehat{F}}$ & -8.48 & \textbf{-28.69} & -4.31 & -9.77 \\
 & $\ln P_{\widetilde{F}}$ & -39.57 & -37.74 & \textbf{-77.57} & -36.40 \\ \hline
\end{tabular}
}
\label{xai_architecture}
\end{table}

\subsubsection{Evaluation on Different NN Architectures}
Apart from evaluation on different datasets, we do experiments on different neural network architectures for CIFAR10 dataset. Results in Table~\ref{xai_architecture} shows that Integrated Gradients maintain the most robust XAI method to misinterpretation with preserved classification, invariant to the change of neural network architecture. However, the robustness to misinterpretation conditioned on misclassification varies according to the internal structure of neural network. GradCAM seems to be robust in most cases. 

\subsubsection{Evaluation for Real-world Models}

\begin{table}[!htbp]
\centering
\caption{Robustness evaluation for Wide ResNet-50-2 model trained on ImageNet dataset. Results are averaged over 20 samples.}
\resizebox{\linewidth}{!}{
\begin{tabular}{ccccc}
\hline
\multirow{2}{*}{XAI Methods} & \multicolumn{2}{c}{Worst Case Evaluation} & \multicolumn{2}{c}{Probabilistic Evaluation} \\ \cline{2-5} 
 & \begin{tabular}[c]{@{}c@{}}$sol_{\widehat{F}}$\\ PCC\end{tabular} & \begin{tabular}[c]{@{}c@{}}$sol_{\widetilde{F}}$\\ (PCC)\end{tabular} & $\ln P_{\widehat{F}}$ & $\ln P_{\widetilde{F}}$ \\ \hline
Gradient x Input & 0.159 & 0.463 & -4.595 & -100 \\
Integrated Gradients & 0.191 & 0.515 & -39.235 & -100 \\
GradCAM & 0.233 & 0.944 & -98.725 & -76.688 \\
FullGrad & 0.315 & 0.799 & -100 &  -75.716 \\
Extremal Perturbations & 0.126 & 0.957 & -4.321 &  -32.612\\ \hline
Accuracy & \multicolumn{2}{c}{Top-1: 81.60\%} & \multicolumn{2}{c}{Top-5: 95.76\%} \\ \hline
\end{tabular}
}
\label{xai_new}
\end{table}

We add additional experiments on wide\_ResNet50\_2 model trained on ImageNet-1K dataset in Table~\ref{xai_new}. We discover that FullGrad aggregates layer-wise gradient maps and thus combine the advantages of Gradient x Input and GradCAM. Extremal Perturbations seek to find the region of an input image that maximally excites a certain output, which is not robust to the adversarial perturbation.

\end{document}